\pdfoutput=1

\documentclass[11pt]{article}

\usepackage[]{acl}

\usepackage{times}
\usepackage{latexsym}

\usepackage[T1]{fontenc}

\usepackage[utf8]{inputenc}

\usepackage[bulgarian,english]{babel}

\usepackage{microtype}

\usepackage{booktabs}
\usepackage{graphicx}
\usepackage{enumitem}
\usepackage{multirow}
\usepackage{makecell}
\usepackage{xspace}
\usepackage{todonotes}
\usepackage{amsmath}
\usepackage{amsfonts}
\usepackage{tabularx}
\usepackage{pifont}
\usepackage{tabularx}
\usepackage{bbding}
\usepackage{fontawesome}
\usepackage{caption}
\usepackage{subcaption}

\newcommand\benchmarkName{bgGLUE\xspace}
\newcommand\benchmarkURL{\url{https://bgglue.github.io}}

\newcommand\PaperTitle{\emph{\benchmarkName}: {A} {B}ulgarian General Language Understanding \\ Evaluation Benchmark}

\newcommand{\xmark}{\ding{55}}%

\title{\PaperTitle}

\author{Momchil Hardalov\thanks{$^*$Work done while Momchil was in the Sofia University, prior to joining Amazon.} \\
  AWS AI Labs \\
  \And
  Pepa Atanasova  \\
  University of Copenhagen, DIKU \\
  \And
  Todor Mihaylov  \\
  Meta AI \\
  \AND
  Galia Angelova \hspace{2cm} Kiril Simov \hspace{2cm} Petya Osenova \\
  IICT, Bulgarian Academy of Sciences \\
  \AND
  Ves Stoyanov  \\
   Meta AI \\
  \And
  Ivan Koychev \\
  Sofia University \\
  \And
  Preslav Nakov \\
  MBZUAI \\
  \And
  Dragomir Radev \\
  Yale University\\
}

\begin{document}
\maketitle
\begin{abstract}

We present \benchmarkName (Bulgarian General Language Understanding Evaluation), a benchmark for evaluating language models on Natural Language Understanding (NLU) tasks in Bulgarian. 
Our benchmark includes NLU tasks targeting a variety of NLP problems (e.g., natural language inference, fact-checking, named entity recognition, sentiment analysis, question answering, etc.) and machine learning tasks (sequence labeling, document-level classification, and regression). We run the first systematic evaluation of pre-trained language models for Bulgarian, comparing and contrasting results across the nine tasks in the benchmark. The evaluation results show strong performance on sequence labeling tasks, but there is a lot of room for improvement for tasks that require more complex reasoning. We make \benchmarkName publicly available together with the fine-tuning and the evaluation code, as well as a public leaderboard at \benchmarkURL, and we hope that it will enable further advancements in developing NLU models for Bulgarian.
\end{abstract}

\section{Introduction}

Natural Language Understanding (NLU) benchmarks, such as GLUE~\cite{wang-etal-2019-glue} and SuperGLUE~\cite{neurips2019_4496bf24}, were designed for a rigorous evaluation of language models on a diverse set of natural language understanding (NLU) tasks. The wide adoption of such benchmarks has driven the rapid development of models that perform well on the tasks that are part of these benchmarks, but also beyond~\cite{devlin2019bert,liu2019roberta}. However, until recently, the focus of such benchmarks has been on English, with little interest in other languages~\cite{bender2011achieving,ponti2019modeling}. 

To address this, recent work has designed benchmarks to test models on NLU tasks on non-English languages~\cite{le-etal-2020-flaubert-unsupervised,rodriguez2021catalan,shavrina-etal-2020-russiansuperglue} or on multiple languages~\cite{liang2020xglue,pmlr-v119-hu20b}.

Here, we aim to improve the diversity of the languages represented in NLU benchmarks by proposing \emph{\benchmarkName}, a benchmark for Bulgarian that consists of nine NLU tasks, including token classification, regression, and classification. Thus far, only individual datasets in Bulgarian have been used for model development and evaluation. Small subsets of up to three downstream tasks in Bulgarian have also been included in some multilingual benchmarks~\cite{liang2020xglue,pmlr-v119-hu20b}. Additionally, there are existing benchmarks focusing on other Balto-Slavic languages, such as the Russian SuperGLUE~\cite{shavrina-etal-2020-russiansuperglue} and the Slovene SuperGLUE~\cite{zagar-robnik-sikonja-2022-slovene}. However, there are no comprehensive benchmarks for representatives of the Eastern South-Slavic language subgroup, and for Bulgarian in particular. We aim to address these limitations with \benchmarkName.

\benchmarkName unifies and facilitates access to existing datasets and tasks for Bulgarian. By including more challenging tasks such as natural language inference, fact-checking, and question answering, we ensure that it comprises a rigorous testset for NLP models developed for Bulgarian. We also provide access to the benchmark through the HuggingFace Hub~\cite{lhoest-etal-2021-datasets} to allow for ease of use and we encourage model sharing for Bulgarian. Moreover, we fine-tune and we run the first systematic evaluation of existing language models for Bulgarian, comparing and contrasting results across all tasks in the benchmark. Our evaluation results show that larger and more robustly pre-trained models yield better performance on all tasks, but also that efficiently distilled models are a strong competitor to their larger counterparts.

\begin{table*}[t]
    \centering
    \resizebox{\textwidth}{!}{%
    \begin{tabular}{llrrrcccl}
        \toprule
         \bf{\#} & \bf{Corpus} & \bf{$|$Train$|$} & \bf{$|$Dev$|$} & \bf{$|$Test$|$} & \bf{Splits} & \bf{Task} &  \bf{Metrics} & \bf{Domain}  \\
         \midrule
         \multicolumn{8}{c}{Token Classification} \\
         \midrule
         1 & \textbf{BSNLP} & 724 & 301 & 301 & \faRefresh & Named Entities & Macro F1 & Misc. \\
         2 & \textbf{PAN-X} & 16,237 & 7,029 & 7,263 & \faSearchMinus & Named Entities & Macro F1 & Wikipedia \\
         3 & \textbf{U.Dep} & 8,907 & 1,115 & 1,116 & & POS Tagging & Macro F1 & Misc. \\
         \midrule
         \multicolumn{8}{c}{Regression / Ranking} \\
         \midrule
         4 & \textbf{Cinexio} & 8,155 & 811 & 861 & \faRefresh~\faSearchMinus & Sentiment & Pear./Spear. Corr. & \makecell[l]{Movies} \\
         5 & \textbf{CT21.T1}  & 2,995 & 350 & 357 & & Check-Worthiness &	Avg. Precision  & Tweets \\
         \midrule
         \multicolumn{8}{c}{Classification Tasks} \\
         \midrule
         6 & \textbf{Cred.-N}  & 19,227 & 5,949 & 17,887 & \faRefresh\xspace~\faPlus & Humor Detection &	Binary F1  & News \\
         7 & \textbf{Fake-N}  & 1,990 & 221 & 701 & \faRefresh & Fake News & Binary F1  & News \\
         8 & \textbf{XNLI} & 392,702 & 5,010 & 2,490 & & NLI & Accuracy & Misc. \\
         9 & \textbf{EXAMS}  & 1,512 & 365 & 1,472 & \faPlus & Multi-Choice QA &	Accuracy & \makecell[cl]{\emph{HS} Exams} \\
         \bottomrule
    \end{tabular}
    }
    \caption{Summary of the tasks included in the \emph{\benchmarkName} benchmark. The numbers in the train, development, and test columns are in terms of examples. The following columns define the structure of the tasks. The domain is based on the source of the texts. The \emph{EXAMS} dataset is collected from high school (HS) examinations. Splits: \faRefresh~new splits; \faSearchMinus~removed duplicates; \faPlus~new examples added/collected.}
    \label{tab:tasks_dataset}
\end{table*}

The models show strong performance on part-of-speech tagging and named entity recognition, but struggle on tasks that require more complex reasoning such as solving matriculation exams, or evaluating the credibility and the veracity of news articles. 
Our contributions are as follows:

\begin{itemize}
    \item We propose the first benchmark for evaluating the capabilities of language models on NLU in Bulgarian, \benchmarkName, which includes nine diverse and challenging downstream tasks.\footnote{The \benchmarkName code, data, and models are available at \\\url{https://github.com/bgGLUE/bgglue}.}
    \item While creating the benchmark, we curated the datasets and created standard splits, where those have not been previously available in the original publications. This facilitates the principled evaluation of all datasets in \benchmarkName.
    \item We train and share 36 models for Bulgarian and provide the first comparative evaluation of existing models on all tasks in \benchmarkName.
\end{itemize}
\section{Tasks}
\label{sec:tasks}

Table~\ref{tab:tasks_dataset} shows the nine datasets that are included in the \benchmarkName benchmark. 
Table~\ref{tab:pertask:examples} shows examples from each dataset and their corresponding labels (translations are available in Table~\ref{tab:appx:en:pertask:examples} in the Appendix). We present additional details such as word overlaps, domain, topic, label distributions, etc. about each dataset in Appendix~\ref{appx:task:details}.

\subsection{Token Classification}

\paragraph{BSNLP} The dataset is released as part of the Balto-Slavic NLP workshop series~\cite{piskorski-etal-2017-first,piskorski-etal-2019-second,piskorski-etal-2021-slav}. The task focuses on cross-lingual document-level extraction of named entities: the systems should recognize, classify, extract, normalize, and make a cross-lingual linking of all named entity mentions in a document; detecting the position of each named entity mention is not required. The target tags are person (PER), organization (ORG), location (LOC), product (PRO), and event (EVT).

\paragraph{PAN-X (WikiANN)} The PAN-X dataset~\cite{pan-etal-2017-cross} has Named Entity Recognition (NER) annotations for persons (PER), organizations (ORG), and locations (LOC). It has been constructed using the linked entities in Wikipedia pages for 282 different languages.

\begin{table*}[th!]
\centering \footnotesize
{
\renewcommand{\arraystretch}{1.0}
\begin{tabular}{p{0.005\textwidth}p{0.93\textwidth}}

 \toprule
 \parbox[t]{1mm}{\multirow{2}{*}{\rotatebox[origin=c]{90}{{\textbf{BSNLP}}}}} &
{\textbf{Document}: 
{\textit{... \foreignlanguage{bulgarian}{Канцлерът на \{Германия\}}\textsuperscript{LOC} \foreignlanguage{bulgarian}{\{Ангела Меркел\}}\textsuperscript{PER} \foreignlanguage{bulgarian}{и президентът на \{Русия\}}\textsuperscript{LOC} \foreignlanguage{bulgarian}{\{Владимир Путин\}}\textsuperscript{PER} \foreignlanguage{bulgarian}{са обсъдили по телефона реализацията на проекта ``\{Северен поток - 2\}}\textsuperscript{PRO}'' \foreignlanguage{bulgarian}{... По - рано компанията ``\{Норд стрим\}}\textsuperscript{ORG}'' \foreignlanguage{bulgarian}{, която води строителството ...}}}} \\ 
& \textbf{Possible Tags}: \underline{Person (PER)}, \underline{Organization (ORG)}, \underline{Location (LOC)}, \underline{Product (PRO)}, \underline{Event (EVT)} \\

\midrule
\parbox[t]{1mm}{\multirow{2}{*}{\rotatebox[origin=c]{90}{{\textbf{Cinexio}}}}} &
{\textbf{User Review}: 
\foreignlanguage{bulgarian}{\textit{Пет звезди са му малко - заслужава поне още толкова :)}}} \\ 
& \textbf{Rating}: \underline{5.0} \\[1em]

\midrule
\parbox[t]{1mm}{\multirow{3}{*}{\rotatebox[origin=c]{90}{{\textbf{Cred.-N}}}}} & \textbf{Body:} 
\foreignlanguage{bulgarian}{\textit{Днес изтича срокът, в който българите, живеещи в чужбина, могат да подадат заявление за разкриване на изборна секция за предстоящия на 27 януари референдум. Според решение на Централната избирателна комисия (ЦИК) за допитването секции могат да се откриват в посолствата и консулствата на страната. За целта обаче са нужни поне 20 заявления на желаещи...}} \\
& \textbf{Title}: 
\foreignlanguage{bulgarian}{\textit{Днес изтича срокът за подаване на заявления за разкриване на секции в чужбина за референдума}} \\
& \textbf{Correct Label}: \underline{Credible}\\

\midrule
\parbox[t]{1mm}{\multirow{2}{*}{\rotatebox[origin=c]{90}{{\textbf{CT21.T1}}}}} &
\textbf{Tweet:} 
\foreignlanguage{bulgarian}{\textit{Според изследване, \#COVID19 оцелява до 3 часа в аерозоли във въздуха, до 24 часа на хартиена и около 2-3 дни на стоманена или пластмасова повърхност. [URL]}}\\
& \textbf{Check-worthy}: \underline{Yes}\\
[0.5em]

\midrule
\parbox[t]{1mm}{\multirow{2}{*}{\rotatebox[origin=c]{90}{{\textbf{EXAMS}}}}} &
\textbf{Paragraph:} 
\foreignlanguage{bulgarian}{\textit{През есента на 917 година той изпраща армия ... за да нападнат Сърбия и да накажат Гойникович за предателството му. Българският владетел отново изпраща Теодор Сигрица и Мармаис, но този път те претърпяват поражение... което принуждава Симеон да сключи примирие с Византия...}} \quad \textbf{Subject}: \textit{History} \\
&  \textbf{Question:} 
\foreignlanguage{bulgarian}{\textit{Кои пълководци оглавяват наказателния поход на Симеон срещу възникналата сръбска опасност през 917 г.?}} \\
& \textbf{Candidate answers:} \\
& 
\textit{(\texttt{A}) 
\foreignlanguage{bulgarian}{\underline{Теодор Сигрица и Мармаис}}}, 
\textit{(\texttt{B}) 
\foreignlanguage{bulgarian}{Кракра и Алусиан}}, 
\textit{(\texttt{C}) 
\foreignlanguage{bulgarian}{Ивац и Никулица}}, 
\textit{(\texttt{D}) 
\foreignlanguage{bulgarian}{Книн, Имник и Ицвоклий}}  \\

\midrule
\parbox[t]{1mm}{\multirow{2}{*}{\rotatebox[origin=c]{90}{{\textbf{Fake.-N}}}}} & 
\textbf{Body:} 
\foreignlanguage{bulgarian}{\textit{Изследователят на българските пророци Христо Радев разкрива предсказания на феномена Слава Севрюкова в интервю за ,,България днес'' ,,В края на 80-те години Слава Севрюкова казва, че в България изневиделица ще се появи човек, в който е прероден духът на ярък библейски герой. Има предвид Давид. Според ясновидката този българин ще изпълни много важна роля в бъдещето на страната. Дано този президент да е въпросният човек! Румен Радев изскочи от нищото, също като библейския Давид...}} \\
& \textbf{Title}: 
\foreignlanguage{bulgarian}{\textit{Petel.bg - новини - ,,България днес'': Изкопаха изгубеното пророчество на Слава Севрюкова за България! То се сбъдва пред очите ни}} \\
& \textbf{Correct Label}: \underline{Fake}\\
\midrule
\parbox[t]{1mm}{\multirow{2}{*}{\rotatebox[origin=c]{90}{{\textbf{PAN-X}}}}} &
\textbf{Sentence:} 
\textit{\foreignlanguage{bulgarian}{Видът е разпространен в \{Бурунди\}}\textsuperscript{LOC}, \foreignlanguage{bulgarian}{\{Демократична република Конго\}}\textsuperscript{LOC}, \foreignlanguage{bulgarian}{\{Замбия\}}\textsuperscript{LOC} \foreignlanguage{bulgarian}{и \{Танзания\}}\textsuperscript{LOC}.} \\
& \textbf{Possible Tags}: \underline{Person (PER)}, \underline{Organization (ORG)}, \underline{Location (LOC)} \\
\midrule
\parbox[t]{1mm}{\multirow{1}{*}{\rotatebox[origin=c]{90}{{\textbf{U.Dep}}}}} & 
\textbf{Sentence:} 
\textit{
\foreignlanguage{bulgarian}{В}\textsuperscript{ADP} \foreignlanguage{bulgarian}{дискусията}\textsuperscript{NOUN} ,\textsuperscript{PUNCT} \foreignlanguage{bulgarian}{предполагам}\textsuperscript{VERB} ,\textsuperscript{PUNCT} \foreignlanguage{bulgarian}{ще}\textsuperscript{AUX} \foreignlanguage{bulgarian}{се}\textsuperscript{PRON} \foreignlanguage{bulgarian}{засегнат}\textsuperscript{VERB} \foreignlanguage{bulgarian}{важни}\textsuperscript{ADJ} \foreignlanguage{bulgarian}{въпроси}\textsuperscript{NOUN} .\textsuperscript{PUNCT}} \\
& \textbf{Possible Tags}: \underline{NOUN}, \underline{PUNCT}, \underline{ADP}, \underline{VERB}, \underline{ADJ}, \underline{PRON}, \underline{AUX}, \underline{PROPN}, \underline{ADV}, \underline{CCONJ}, \underline{DET}, \underline{NUM}, \underline{PART}, \underline{SCONJ}, \underline{INTJ} \\[1em]

\midrule
\parbox[t]{1mm}{\multirow{2}{*}{\rotatebox[origin=c]{90}{{\textbf{XNLI}}}}} &
\textbf{Text:} 
\foreignlanguage{bulgarian}{\textit{И той каза: Мамо, у дома съм. Той се обади на майка си веднага щом училищният автобус го е оставил.}}\\
& \textbf{Hypothesis:} 
\foreignlanguage{bulgarian}{
\foreignlanguage{bulgarian}{\textit{Той се обади на майка си веднага щом училищният автобус го е оставил.}}} \\ 
& \textbf{Entailment:} \underline{Neutral}\\

\bottomrule
\end{tabular}
}
\caption{Examples from our \benchmarkName benchmark. For each task, the different parts of the example are shown in \textbf{Bold}. 
\underline{Underlined} text shows the label for that example (or the set of possible labels). The precise model's inputs and the expected outputs (labels) are shown in Table~\ref{tab:input:output} in the Appendix. Translations for each examples are shown in Table~\ref{tab:appx:en:pertask:examples}.}
    \label{tab:pertask:examples}
\end{table*}

\paragraph{Universal Dependencies (U. Dep)} Universal Dependencies (UD)~\cite{nivre-etal-2020-universal} is a framework for consistent annotation of grammar (part of speech, morphological features, and syntactic dependencies) across different human languages. UD is an open community effort with over 300 contributors producing more than 200 treebanks in over 100 languages. The dataset was collected, annotated, and later transferred into the required UD format as part of the Bulgarian treebank (BTB-BulTreeBank) project \cite{osenova-simov-2015-universalizing}.

\subsection{Natural Language Inference}
\paragraph{XNLI} This dataset~\cite{conneau-etal-2018-xnli} is a subset of a few thousand examples from MNLI, which has been translated into 14  languages. As with MNLI, the goal is to predict textual entailment: does sentence A imply/contradict/neither sentence B. This is a classification task: given two sentences, predict one of the three labels.

\subsection{Sentiment Analysis}
\paragraph{Cinexio} The Cinexio dataset~\cite{kapukaranov-nakov-2015-fine} focuses on fine-grained sentiment analysis of movie reviews. It was automatically collected to contain movie reviews in Bulgarian from the Cinexio ticket-booking website (which is not available anymore).

\subsection{News Credibility / Fact-Checking}
\paragraph{CLEF-2021 CheckThat!, Task 1A (CT21.T1)} Check-Worthiness Estimation dataset is part of the 2021 CheckThat! Lab on Detecting Check-Worthy Claims, previously Fact-Checked Claims, and Fake News (Task 1)~\cite{clef-checkthat:2021:task1}. The aim of the task is to determine whether a piece of text is worth fact-checking. More precisely, given a tweet, one has to produce a ranked list of tweets, ordered by their check-worthiness.

\paragraph{Credible News (Cred.-N)} The \emph{Credible News}~\cite{10.1007/978-3-319-44748-3_17} dataset focuses on the problem of automatically distinguishing credible from fake and humorous news. The examples are articles collected from six Bulgarian news websites. The articles cover various topics including politics (both local and global), sports, lifestyle, and pop culture. The original dataset contained news from four websites. As part of the \benchmarkName initiative, we collected 6,550 new articles (5K credible and 1.5K humorous) from two new websites, and we release more than 30K ones that were not publicly available.

\paragraph{Fake News (Fake-N)} This dataset~\cite{hackthefakenews2017,karadzhov-etal-2017-built} contains Bulgarian news articles over a fixed period of time, whose factuality was questioned. These news articles come from 377 different sources from various domains, including politics, interesting facts, and tips\&tricks. The dataset was prepared for and used in the \emph{Hack the Fake News} hackathon in 2017. We found and removed instances that were duplicated across the splits and we further randomly allocated 10\% of the training instances for a development dataset, which was not available in the original version of the dataset. 

\subsection{Question Answering}
\paragraph{High School Examinations (EXAMS)} EXAMS~\cite{hardalov-etal-2019-beyond,hardalov-etal-2020-exams} is a benchmark dataset for cross-lingual and multilingual question answering for high school examinations. It contains more than 24,000 high-quality exam questions in 26 languages, covering eight language families and 24 school subjects from the Natural Sciences and Social Sciences, among others. EXAMS offers a fine-grained evaluation framework across multiple languages and subjects, which allows a precise analysis and comparison of various models.

\subsection{Scoring} In \benchmarkName, we opt for the simple approach of weighing each task equally, and for tasks with multiple metrics, first averaging those scores to get a task score.
\section{Data Preparation}
\label{sec:data:preparation}

Here, we describe the pre-processing steps we took to prepare the datasets before including them in the \benchmarkName benchmark. Our main goal was to ensure that the setup evaluated the language understanding abilities of the models in a principled way and in a diverse set of domains. Since all of the datasets were publicly available, we preserved the original setup as much as possible. Nevertheless, we found that some datasets contained duplicate examples across their train/dev/test splits, or that all of the splits came from the same domain, which may overestimate the model's performance. Hereby, \textit{we removed data leaks} and \textit{proposed new topic-based or temporal-based (i.e.,~timestamp-based)} data splits where needed. We deduplicated the examples based on a complete word overlap in two pairs of normalized texts, i.e.,~lowercased, and excluding all stop words.

For the \emph{BSNLP} task, we combined the data from three consecutive editions of the NER shared task. We selected all Bulgarian examples, which encompassed six different topics. We used the latest two topics for testing, and split the rest randomly at a 4:1 ratio for training and validation.

The \emph{Credible News} dataset contained data from six sources and various news topics. We split the dataset both by topic and by source. In particular, we included all news articles from the same topic in the same split. For training and validation, we used documents from the largest sources from the two categories. Moreover, we extracted the same topics from the two and grouped them together within the splits, keeping the class ratio at 1:10. In the test dataset, we used data points from all six data sources.\footnote{We did not have overlapping topics from the same source in the different splits.} We note that our test set contains all data points from the two new sources, which are more recent, making them even more challenging. Finally, we cleaned the texts from duplicates and removed all keywords indicating the source: the names of the authors, the media source, the URLs, etc. More details about the collection process of the news articles are given in Appendix~\ref{appx:task:details}.

We prepared an entirely new split for \emph{Cinexio}, as there was no standard one. First, we removed all duplicate comments and we sorted the remaining ones by publication time. Next, we split the comments using half for training and the rest equally for validation and testing. Finally, we removed from the training set the comments for the same movie that appeared in the other two splits, which changed the distribution by slightly increasing the proportion of the training set.

\begin{table*}[th]
    \centering
    \resizebox{\textwidth}{!}{%
    \setlength{\tabcolsep}{3.5pt}
    \begin{tabular}{rl|c|ccccccccc}
    \toprule
     {} &  & \bf{bgGLUE} & \bf{BSNLP} &  \bf{Cinexio} &  \bf{CT21.T1} &  \bf{Cred.-N} &  \bf{EXAMS} &  \bf{Fake-N} &  \bf{U.Dep} &  \bf{PAN-X} &  \bf{XNLI} \\
      \bf{\#} & \bf{Model Name} & Avg. $\xrightarrow{}$  & F1\textsubscript{macro} & P/S Corr. & Avg. P & F1\textsubscript{binary} & Acc. & F1\textsubscript{binary} & F1\textsubscript{macro} & F1\textsubscript{macro} & Acc. \\
    \midrule
        \multicolumn{12}{c}{\bf{Random Baselines}} \\
        \midrule
         - &  Majority &  18.52 &   0.00 &     0.00 &    28.41 &    34.14 &  25.68 &   45.08 &   0.01 &   0.00 & 33.33 \\
         - &    Random &  17.59 &   0.75 &     0.00 &    25.06 &    30.14 &  25.54 &   35.65 &   6.31 &   0.94 & 33.33 \\
        \midrule
        \multicolumn{12}{c}{\bf{Fine-tuning Baselines}} \\ 
    \midrule
     
         1 &                      XLM-R\textsubscript{large} &  \textbf{75.82} &  \underline{63.81} &    \textbf{85.69} &    \textbf{69.45} &    \textbf{79.73} &  \textbf{36.41} &   \textbf{70.31} &  \textbf{99.30} &  \textbf{92.96} & \textbf{84.71} \\
         2 &                       XLM-R\textsubscript{base} &  \underline{73.04} &  62.47 &    \underline{84.40} &    63.91 &    \underline{75.74} &  33.42 &   66.82 &  \underline{99.23} &  91.18 & \underline{80.22} \\
         3 & SlavicBERT &  72.12 &  $^\ddagger$\textbf{65.28} &    81.71 &    62.70 &    72.01 &  31.86 &   \underline{67.28} &  99.06 &  \underline{92.36} & 76.79 \\
         4 &           mBERT\textsubscript{base} &  71.08 &  56.13 &    82.07 &    64.79 &    69.17 &  \underline{35.39} &   65.65 &  98.99 &  92.11 & 75.39 \\
         5 & MiniLM\textsubscript{L12} &  70.96 &  59.70 &    80.63 &    57.37 &    75.41 &  {35.26} &   64.33 &  98.91 &  90.26 & 76.81 \\
         6 &     Distil-mBERT &  69.58 &  52.82 &    80.32 &    \underline{65.15} &    67.05 &  34.31 &   65.66 &  98.58 &  90.82 & 71.50 \\

    \bottomrule
    \end{tabular}
    }
    \caption{Baseline results on the \benchmarkName benchmark. We show the best results in \textbf{bold} and we \underline{underline} the second best result. The scores for each model are the highest ones achieved during hyper-parameter search by selecting the best model checkpoint on each task's development set. We calculate the \emph{\benchmarkName} score on the raw scores (before rounding) and then we round it to two digits. Following the notation of previous benchmarks, we multiply the results by 100. $^\ddagger$\emph{SlavicBERT} is pre-trained on all languages from the \emph{BSNLP} NER task (not using our splits), therefore its score on that task is unrealistically high.
    }
    \label{tab:baselines}
\end{table*}

For \emph{EXAMS}, we kept the original validation and test splits~\cite{hardalov-etal-2020-exams}. We added all additional questions from \citet{hardalov-etal-2019-beyond} to the training set, i.e.,~the category \emph{history online quizzes}.

The \emph{Fake News} dataset was released as part of a shared task and it was already split into training and testing sets. After manual inspection, we found that some of the articles had only their titles reworded and had the same content. Therefore, we removed duplicates based only on exact matches in the article's body. In addition, we designated 10\% of the training examples for validation.

\emph{PAN-X} contained short sentences from Wikipedia automatically annotated based on the available Wiki entities. The dataset consists of 20K examples for training and 10K for validation and testing. We kept the proposed splits, but we checked for duplicates by converting each sentence to lower case and removing the punctuation. This resulted in the removal of 10K sentences overall.

Our analysis did not find any issues with \emph{CT21.T1}, \emph{U.Dep}, and \emph{XNLI}, and thus we kept the splits as provided originally.

\section{Experiments}
\label{sec:experiments}

In this section, we first describe the baseline systems we experiment with and then we present the evaluation results.

\subsection{Baselines}

\paragraph{Majority and Random Baselines} The majority class baseline is calculated from the distributions of the labels in each test set. In the random baseline, each test instance is assigned a target label at random with equal probability.

\paragraph{Fine-tuned models}

Our baselines include several prominent multilingual 
encoder-only pre-trained Transformer models. We divide them, based on their pre-training objective as follows: 


\begin{enumerate}[label=\Roman*, wide, labelindent=0pt,nosep]
    \item \textit{Masked language modeling}:
    \begin{itemize}
        \item \textbf{mBERT}~\cite{devlin2019bert} We use the \emph{base} cased version, trained on 104 languages, including Bulgarian. The pre-training task is done on a Wikipedia dump for each language.
        \item \textbf{XLM-R}~\cite{conneau2020-xlm-roberta} We evaluate the \emph{Base} and the \emph{Large} versions of the model. They are trained on filtered CommonCrawl data in 100 languages, including Bulgarian.
    \end{itemize}
    \item \textit{Knowledge distillation}:
    \begin{itemize}
        \item \textbf{Distil-mBERT}~\cite{sanh2019distilbert} The model is distilled using mBERT as the teacher.
        \item \textbf{MiniLM\textsubscript{L12}}~\cite{NEURIPS2020_3f5ee243} The model is distilled using XLM-R\textsubscript{Base} as the teacher, on the same pre-training corpora as the latter.
    \end{itemize}
    \item \textit{Bulgarian downstream task}:
    \begin{itemize}
        \item \textbf{SlavicBERT}~\cite{piskorski-etal-2021-slav} The model is based on mBERT that is additionally pre-trained with four Slavic languages: Bulgarian, Czech, Polish, and Russian, using a stratified dataset of Russian news and Wiki articles for the other languages. Finally, the model is fine-tuned on all the languages from the BSNLP shared task.
    \end{itemize}
\end{enumerate}

For the token classification tasks (\emph{BSNLP}, \emph{U.Dep}, \emph{PAN-X}), we predict the tag for each word based on the tag of the first sub-token. For the sentence classification tasks, we obtain the predictions based on the first special token (e.g.,~\emph{[CLS]}). For \emph{Cinexio}, we optimize a mean squared error loss. For \emph{CT21.T1}, \emph{Cred.-N}, \emph{Fake-N}, and \emph{XNLI}, we optimize the cross entropy loss. Finally, for \emph{EXAMS}, we optimize a binary cross entropy for each candidate answer. More details about the experimental setup, the values of the model hyper-parameters, and other training details can be found in Appendix~\ref{appx:model:hyperparams}. For a description of the inputs and the outputs, we refer the reader to Appendix~\ref{appx:model:inputs}.

\subsection{Experimental Results}



Table~\ref{tab:baselines} shows the results for the baseline models fine-tuned on the \emph{\benchmarkName} tasks. Each model is trained on one task at a time. First, we see that the random and the majority baselines achieve below 20 points \emph{\benchmarkName score}, and all fine-tuned models outperform them on all tasks by a sizable margin. 

\begin{figure}
    \centering
    \includegraphics[width=1.04\columnwidth]{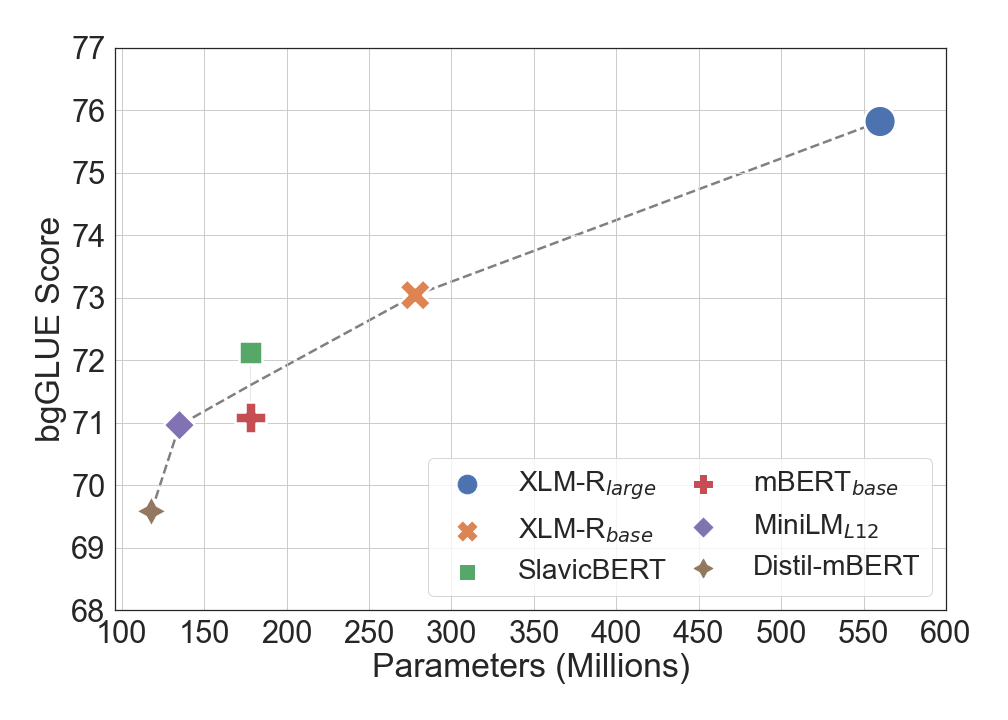}
    \caption{Scaling curve of the \benchmarkName score based on the model size (in million parameters). The dotted line illustrates the average increase.}
    \label{fig:score_scale}
\end{figure}

Figure~\ref{fig:score_scale} shows the correlation between the model size and the \benchmarkName score: we can see that scaling the model size brings additional performance improvements~\cite{devlin2019bert,conneau2020-xlm-roberta,brown2020language,chowdhery2022palm,zhang2022opt,scao2022bloom}.

Table~\ref{tab:model:parameters} in the Appendix summarizes the number of parameters for each model we experimented with. We can see in that there is a linear correlation between the number of parameters in the model and its peformance, which holds for all models  except for Distil-mBERT, where the slope is more steep. Nonetheless, it is clear that the model size is not the only factor that affects the down-stream task performance. Other important factors include the pre-training dataset that was used~\cite{liu2019roberta,raffel2020exploring}, the number of tokens in that training set~\cite{hoffmann2022training}, whether the model is distilled or is full-size~\cite{sanh2019distilbert,NEURIPS2020_3f5ee243}, whether it is monolingual or multilingual~\cite{conneau2020-xlm-roberta}, etc.

As expected, the largest model, XLM-R\textsubscript{Large}, outperformed its smaller version XLM-R\textsubscript{Base} by 2.78 points absolute. Moreover, the more robustly pre-trained XLM-R model on average outperformed by 2 points a similar-sized mBERT, scoring at 73.0 \benchmarkName score. The highest differences between BERT-based and XLM-R-based models (including their distilled versions) are on \emph{BSNLP}, \emph{Cinexio}, \emph{Cred.-N}, and \emph{XNLI}, in favor of XLM; on \emph{CT21.T1} and \emph{PAN-X} this is in favor of BERT.

The gap between XLM-R and mBERT is reduced by 1 point absolute by the SlavicBERT's additional fine-tuning on downstream tasks. Although the largest improvement is observed for NER tasks,\footnote{SlavicBERT is pre-trained on data from the 2019 edition of the BSNLP competition using different splits.} we see an increase by 1-2 points also on \emph{Cred.-N}, \emph{Fake-N}, and \emph{XNLI}, compared to mBERT. However, we also see that the downstream pre-fine-tuning is not beneficial for all tasks~\cite{poth-etal-2021-pre}, and we see a drop in performance for ranking (\emph{CT21.T1}) and question answering\footnote{\citet{hardalov-etal-2019-beyond} observed the same when fine-tuning SlavicBERT for multiple-choice QA.} (\emph{EXAMS}) tasks.

Our evaluation of knowledge-distilled models shows that they are a competitive alternative to their teacher models for Bulgarian. Although they are ranked last in terms of performance, their results are 1.5--2.0 points of \benchmarkName score behind the best results we obtained, and thus we believe they are a viable alternative to the full models. 

In order to measure the trade-off between model size and model performance, we compare mBERT to DistilBERT and XLM-R to MiniLM. In the case of DistilBERT, we have 30\% (178M $\rightarrow$ 135M) fewer parameters, which results in a 2.2\% drop in performance. In turn, MiniLM has less than half of the parameters of its teacher, i.e.,~135\% fewer parameters (278M $\rightarrow$ 118M), which leads to only a 2.9\% relative drop in performance. The most challenging tasks for the distilled models are NER (\emph{BSNLP} and \emph{PAN-X}), NLI (\emph{Cinexio} and \emph{XNLI}), where we see a sizable gap between non-distilled models. Finally, we note that MiniLM has the worst performance on \emph{CT21.T1}, which is also so for XLM-R. We hypothesize that this is due to the source of the dataset being Twitter. For more detailed results, we refer the reader to Appendix~\ref{appx:finegrained:results}.
\section{Discussion}
\label{sec:discussion}
\paragraph{Software Tools} As part of building the benchmark, we developed a set of software tools that facilitate the training and the evaluation of new models. The toolkit is implemented in PyTorch~\cite{paszke2019pytorch}, using the \emph{transformers} library~\cite{wolf-etal-2020-transformers}. Moreover, we integrate and release publicly all datasets (in accordance with their licenses; see the next paragraph) from \emph{\benchmarkName} in the HuggingFace datasets repository~\citep{lhoest-etal-2021-datasets}.\footnote{https://huggingface.co/bgglue}

\paragraph{Dataset Licenses} We keep the licenses as provided by their authors for all datasets included in the \emph{\benchmarkName} benchmark. Table~\ref{tab:licences} summarizes the information about each dataset and gives links to the external websites and code repositories provided by the authors. All datasets are available to use for research purposes. Some of them come with a non-commercial license, i.e.,~\emph{Cinexio}, \emph{Cred.-N}, \emph{U.Dep}, and \emph{XNLI}. \emph{Cred.-N} requires signing an agreement form before obtaining the dataset.

\begin{table}[t]
    \centering
    \resizebox{\columnwidth}{!}{%
    \setlength{\tabcolsep}{3pt}
    \begin{tabular}{lcccc}
    \toprule
        \bf{Task} & \bf{Public} & \bf{License} & \bf{Website} & \bf{Code} \\
    \midrule
        \bf{BSNLP}   & \Checkmark & \xmark & \href{http://bsnlp.cs.helsinki.fi/shared-task.html}{\faLink} & - \\
        \bf{Cinexio} & \Checkmark & \faGraduationCap & \href{http://bkapukaranov.github.io/}{\faLink} & - \\
        \bf{CT21.T1} & \Checkmark & \faGraduationCap & \href{https://sites.google.com/view/clef2021-checkthat/tasks/task-1-check-worthiness-estimation}{\faLink} & \href{https://gitlab.com/checkthat_lab/clef2021-checkthat-lab/-/tree/master/task1}{\faCode} \\
        \bf{Cred.-N} & \xmark  &  \faGraduationCap & - & \href{https://github.com/mhardalov/news-credibility}{\faCode} \\
        \bf{EXAMS}   & \Checkmark & \faCreativeCommons & - & \href{https://github.com/mhardalov/exams-qa}{\faCode} \\
        \bf{Fake-N}  & \Checkmark & \faUnlock & -  & \href{https://gitlab.com/datasciencesociety/case_fake_news}{\faCode} \\
        \bf{U.Dep}   & \Checkmark & \faGraduationCap          & \href{https://universaldependencies.org/}{\faLink} & \href{https://github.com/UniversalDependencies/UD_Bulgarian-BTB}{\faCode} \\
        \bf{PAN-X}   & \Checkmark & \xmark & - & \href{https://github.com/afshinrahimi/mmner}{\faCode} \\
        \bf{XNLI}    & \Checkmark & \faGraduationCap & \href{https://www.nyu.edu/projects/bowman/xnli/}{\faLink} & \href{https://github.com/facebookresearch/XNLI}{\faCode} \\
    \bottomrule
    \end{tabular}
    }
    \caption{Dataset licensing information. The \emph{Cred.-N} dataset is not public and is distributed after filling up an agreement form. All other datasets are publicly distributed under different licenses: \xmark\xspace no specific license, 
    \faGraduationCap\xspace non commercial use only, \faCreativeCommons\xspace creative commons license open for commercial use, \faUnlock\xspace MIT License. \faLink\xspace the dataset has a website. \faCode\xspace the authors offer a repository with code.}
    \label{tab:licences}
\end{table}


\paragraph{Modeling Considerations} 
Previous work has shown that a model's scale~\cite{kaplan2020scaling,hoffmann2022training} is an important factor for its performance~\cite{radford2018gpt1,devlin2019bert,liu2019roberta,radford2019language,conneau2020-xlm-roberta,soltan2022alexatm,zhang2022opt}, especially in zero-shot or few-shot settings~\cite{brown2020language,wei2021finetuned,chowdhery2022palm,NEURIPS2022_b1efde53}. 

Nevertheless, these models are often pre-trained on high-resource languages such as English. A few noteworthy alternatives for Bulgarian include XLM-RoBERTa~\cite{goyal-etal-2021-larger}, multilingual T5~\cite{xue-etal-2021-mt5}, XGLM~\cite{lin-etal-2022-shot}, mGPT~\cite{shliazhko2022mgpt}, and the extended version of BLOOM~\cite{yong2022bloomplus}. These models represent different variants of the Transformer architecture, i.e.,~encoder/decoder-only or sequence-to-sequence. In this work, we only included encoder-only Transformer models with less than one billion parameters (see Table~\ref{tab:model:parameters} in the Appendix). We leave the rest to be explored by future participants in the \benchmarkName benchmark and we note that some of the tasks, such as ranking or regression, require additional steps to make the different architectures work.

Although a model's scale is important for its performance, it comes with additional efficiency and computational costs, among other  considerations~\cite{bommasani2021opportunities}. Fine-tuning large pre-trained language models is usually time-consuming and expensive, and it also requires a large number of manually annotated examples. A possible direction that alleviates these requirements is to use adapter-based models~\cite{NIPS2017_e7b24b11,pfeiffer-etal-2021-adapterfusion} and other techniques for efficient training~\cite{lester-etal-2021-power,hu2021lora,ben-zaken-etal-2022-bitfit}. 

Promising results were shown both in multilingual~\cite{pfeiffer-etal-2020-mad} and in cross-lingual settings~\cite{muennighoff2022crosslingual}. 

Another line of work, which we leave for future research, is zero-shot and few-shot learning. Recently, different techniques have been developed such as learning from demonstrations~\citet{brown2020language}, patterns~\cite{schick-schutze-2021-shot,schick-schutze-2021-just}, instructions~\cite{mishra-etal-2022-cross,wang-etal-2022-super,chung2022scaling,iyer2022opt}, or multi-task fine-tuning~\citet{raffel2020exploring,wei2021finetuned,chowdhery2022palm}. These models require fewer examples due to their extensive fine-tuning, but they still showed \citet{chowdhery2022palm} that a model's size is of crucial importance for their performance.

Finally, there is a trade-off between performance and using monolingual vs. multilingual models~\cite{devlin2019bert, conneau2020-xlm-roberta,pyysalo-etal-2021-wikibert}. Extensively pre-trained monolingual models on language-specific corpora often achieve better performance compared to multilingual ones~\cite{kuratov2019adaptation,canete2020spanish,masala-etal-2020-robert,delobelle-etal-2020-robbert,chan-etal-2020-germans,martin-etal-2020-camembert,cui2021pre,pyysalo-etal-2021-wikibert,barry-etal-2022-gabert}. However, there is no open-source large scale pre-trained monolingual Bulgarian model with extensive pre-training: most of the existing checkpoints are based on multilingual ones, and they are fine-tuned on small corpora.\footnote{Reference: \url{https://huggingface.co/models?language=bg}}

\paragraph{Leaderboard} We develop our leaderboard in accordance with existing ones, e.g.,~the (Super)GLUE~\cite{wang-etal-2019-glue,neurips2019_4496bf24}: the participants are provided with all the training, validation, and test examples without the gold test labels. They submit an archive with their predictions for each task, and then our system automatically evaluates their predictions.

The intended use of our leaderboard is to provide a standardized way to compare the performance of different models on specific tasks, thus allowing researchers and practitioners to assess the current state of the art and to identify areas where improvements can be made. We urge against making improperly supported claims about general language understanding based on the performance on our leaderboard, and on NLP leaderboards in general~\cite{ethayarajh-jurafsky-2020-utility,raji2021ai,blasi-etal-2022-systematic}.

We believe that the \benchmarkName leaderboard will incentivize model and resource creation in two ways: (\emph{i})~the participants are required to share details about their submissions, and are encouraged to release their models; we cannot force the latter, but we can ensure that the methods are reproducible to some extent; (\emph{ii}) practice shows that the results on such leaderboards tend to saturate in several years, which will likely happen with this benchmark as well. We plan to open our platform and to work with interested researchers, first to design new leaderboards~\cite{ma2021dynaboard}, second to include their datasets into \benchmarkName, and third to collaborate to build new (including human-and-model-in-the-loop~\cite{kiela-etal-2021-dynabench}) and refining excising language resources for Bulgarian.








%


%

%

%
\section{Related Work}

\paragraph{Language Understanding}
The release of the code and English corpora as part of the General Language Understanding Evaluation (GLUE~\citet{wang-etal-2019-glue}) was a push towards the development of models with improved performance on a diverse set of downstream tasks. The GLUE benchmark includes 11 NLU tasks, such as semantic textual similarity, natural language inference, and other classification tasks. Later, the benchmark was extended with additional and more sophisticated tasks in its SuperGLUE~\citep{neurips2019_4496bf24} variant. 

While GLUE and SuperGLUE have been established as the de-facto standard for evaluating machine learning models, they are limited to English. To foster the evaluation and the development of machine learning models for other languages, several benchmarks in other languages have been released. They can be grouped based on their language family as follows: \emph{Romance} -- French~\cite{le-etal-2020-flaubert-unsupervised}, Catalan~\cite{rodriguez2021catalan}, \emph{Balto-Slavic} -- Russian~\cite{shavrina-etal-2020-russiansuperglue}, 
Slovenian~\cite{zagar-robnik-sikonja-2022-slovene}, \emph{Iranian} -- Persian~\cite{khashabi-etal-2021-parsinlu}, \emph{Altic} -- Korean~\cite{park2021klue}, \emph{Sino-Tibetan} -- both CLUE~\cite{xu-etal-2020-clue}, and CUGE~\citep{yao2021cuge} focus on Chinese, \emph{Indic} --  \citet{kakwani-etal-2020-indicnlpsuite} evaluated fine-tuned pre-trained models on multiple Indic languages, while \citet{doddapaneni2022indicxtreme} focused on their zero-shot capabilities, and \emph{Malayic} -- Indonesian~\citep{koto-etal-2020-indolem}. \citet{khanuja-etal-2020-gluecos} provides further resources for code-switched languages (English with Spanish or Hindi). 

While \citet{shavrina-etal-2020-russiansuperglue} provided resources for Balto-Slavic languages, there are \textit{no existing benchmarks for languages in the South Eastern-Slavic subgroup or for Bulgarian in particular}. We address this deficiency by developing \emph{\benchmarkName}, a benchmark for Bulgarian, which is part of the South-Slavic subgroup. \citet{10.2478/cait-2021-0027} published a survey of the language resources available for Bulgarian, which we also include in the current benchmark, extended with more recent datasets.

The aforementioned benchmarks focus on a single language or on a single language family. Other studies looked at multiple languages. \citet{liang2020xglue} proposed XGLUE, a benchmark for 19 languages that covers NLU problems and language generation tasks. \citet{pmlr-v119-hu20b} collected a cross-lingual evaluation dataset in 40 languages, later extended with 10 additional \cite{ruder-etal-2021-xtreme}, including tasks similar to the original (Super)GLUE setup such as token classification, question answering, textual similarity, natural language inference, etc. Both benchmarks include Bulgarian, but are limited to three tasks: part-of-speech (POS) tagging (Universal Dependencies~\citet{nivre-etal-2020-universal}), named entity recognition (PAN-X/WikiAnn~\citet{pan-etal-2017-cross}), and natural language inference (XNLI~\citet{conneau-etal-2018-xnli}). These tasks are also part of \emph{\benchmarkName}, but we extend them with additional NLU tasks, including question answering, fake news detection, sentiment analysis, etc.

More recently, a large-scale initiative for providing open-access to large language models trained to perform new tasks based on few demonstrations or natural language instructions was launched as part of the BLOOM workshop~\cite{scao2022bloom}. This led to the release of a new corpus, comprising sources in 46 natural and 13 programming languages, and a multilingual decoder-only Transformer language model pre-trained on that data. However, BLOOM was not pre-trained on Slavic languages, and it was only later that zero-shot support for Bulgarian was added~\citet{yong2022bloomplus}. 

BIG-Bench~\cite{srivastava2023beyond} is another such initiative that incorporates more than 200 tasks (some not related to NLP) to test the capabilities of language models. The task topics are diverse, drawing problems from linguistics, childhood development, math, common-sense reasoning, biology, physics, social bias, software development, and beyond. Currently, there are few non-English tasks included, but none of them is for Bulgarian.

The \emph{low number of Bulgarian resources} that are part of these initiatives is yet another reason why \textit{more publicly available Bulgarian resources and open-access models are needed}.

\paragraph{Other Modalities} Existing work also quantifies the abilities of state-of-the-art models in multimodal settings. CodeX GLUE~\cite{lu2021codexglue} is a benchmark for program understanding and generation. \citet{conneau2022xtremes} proposed XTREME-S that focuses on speech tasks, including  speech recognition, classification, speech-to-text translation, and retrieval. Finally, IGLUE~\cite{pmlr-v162-bugliarello22a} fills the gap in image and text evaluation, including tasks such as visual question answering, cross-modal retrieval, and grounded reasoning. Bulgarian is included as part of both XTREME-S and IGLUE. However, here we focus only on NLP tasks on text and currently we do not include tasks with multiple modalities in the present benchmark.
\section{Conclusion and Future Work}
We presented \emph{\benchmarkName} -- the first holistic benchmark for evaluating NLU systems in Bulgarian. It includes nine challenging tasks that cover token classification, regression/ranking, and text classification. We fine-tuned and evaluated six different pre-trained state-of-the-art language models. Our extensive evaluation showed that \benchmarkName contains challenging tasks that are far from being solved. Finally, we open-sourced the cleaned versions of the datasets, including the new, more challenging splits, and the source code for training and evaluation, and we released 36 fine-tuned models (one for every task and model combination). All the released artifacts are also integrated into the HugginFace Hub. We believe that \benchmarkName is a rich and challenging testbed that will cultivate prospective work on Bulgarian language understanding.

In future work, we plan to add more tasks for Bulgarian, e.g., \cite{dinkov-etal-2019-detecting}.
We also want to use monolingual Bulgarian datasets for pretraining, beyond Wikipedia and CommonCrawl, e.g., \cite{simov2002hpsg,simov-etal-2004-language,koeva2004bulgarian,koeva2012bulgarian,koeva-etal-2020-natural}, using which will require a thorough assessment in order to prevent introducing unwanted biases and hazardous behavior in the models trained on them~\cite{10.1145/3442188.3445922,liang2021towards}. Finally, we plan to try recent multilingual models such as mDeBERTaV3 \cite{he2021debertav3}, mT0 and BLOOMz \cite{muennighoff2023crosslingual}.

\section*{In Memory of Professor Dragomir Radev}

We dedicate this work to the memory of Dragomir Radev, who is a co-author of this paper. Drago had a tremendous impact on our community, and his legacy will live on through the countless students and colleagues whose lives he touched. Drago was not only an exceptional computer scientist, but one of the kindest and most humble people many of us have ever known. He deeply cared about Bulgarian NLP and Bulgarian NLP researchers. He was also the one who gave the idea and who remained the main driving force behind the Bulgarian GLUE project. Drago will be greatly missed... 

\section*{Acknowledgements}

We thank the anonymous reviewers for their helpful questions and comments, which have helped us improve the quality of the paper. We also want to thank the MSc students from Sofia University ``St. Kliment Ohridski'' Mihail Mihaylov and Kiril Georgiev for their help with the initial parsing and analysis. We further thank Quỳnh Mihaylova for designing the \benchmarkName logo and for her priceless feedback on the design of the leaderboard.

This work is partially supported by Project UNITe BG05M2OP001-1.001-0004 funded by the Bulgarian OP ``Science and Education for Smart Growth'', co-funded by the EU via the ESI Funds.

The work on BulTreeBank-UD and the Balto-Slavic datasets for Bulgarian is partially supported by CLaDA-BG, the Bulgarian National Interdisciplinary Research e-Infrastructure for Resources and Technologies in favor of the Bulgarian Language and Cultural Heritage, part of the EU infrastructures CLARIN and DARIAH, grant number DO1-301/17.12.21.
\section*{Limitations} 

\paragraph{Tasks in \benchmarkName} The \benchmarkName benchmark is comprised of  nine challenging NLU tasks, including three token classification tasks, one ranking task and five text classification tasks. While we cover three different types of tasks in the benchmark, we are restricted by the available resources for Bulgarian, and thus we could not include some other NLP tasks, such as language generation. We also consider only NLP tasks and we do not include tasks with other/multiple modalities. Finally, some of the tasks are of similar nature, e.g.,~we include two datasets for NER and two for credibility/fake news classification (see Section~\ref{sec:tasks}).

\paragraph{Domains in \benchmarkName} The tasks included in \benchmarkName span over multiple domains such as social media posts, Wikipedia, and news articles and can test both for short and long document understanding. However, each task is limited to one domain and the topics within the domain do not necessarily have full coverage of all possible topics. Moreover, some of the tasks have overlapping domains, e.g.,~the documents in both Cred.-N and Fake-N are news articles.

\paragraph{Baseline Models}
As described in Section~\ref{sec:discussion}, the baseline models provided for \benchmarkName include fairly small encoder-only Transformer architectures. We leave for future work other modeling architectures and modeling techniques that are known for improving the efficiency and the computational requirements of the used models, e.g.,~few-shot and zero-shot in-context learning and instruction-based evaluation, multi-task learning, etc. 

\paragraph{Model Biases} In this work, we did not explore whether the datasets in \benchmarkName contain unwanted biases, which could also lead to potential hazardous behavior of the baselines we trained in our experiments with the \benchmarkName benchmark.

\section*{Ethics and Broader Impact}

\subsection*{Dataset Collection}

In \benchmarkName, we include only datasets that are publicly available, with a license that allows at a minimum free use for academic research. We have also referenced the original work where the corresponding resources were first proposed. We encourage the users of \benchmarkName to refer to the original work for licensing details.

Additionally, we carefully examined and removed the instances of the dataset that were duplicated across the training/development/test splits. Whenever development or other dataset splits were not available, we also provide new dataset splits as well. Section~\ref{sec:tasks} points where such changes of the corresponding original resources were required, and the code used to filter or to produce the new splits is available in \benchmarkName's code repository. We believe that the selection of publicly available datasets and the adopted dataset curation steps will foster the development and the rigorous evaluation of language models for Bulgarian.

\subsection*{Biases and Misuse Potential}

The datasets included in \benchmarkName were annotated by human annotators, who could be subject to potential biases in their annotation process. Hence, the datasets in \benchmarkName could potentially be misused to develop models that make predictions that are unfair to individuals or groups. Therefore, we ask users of \benchmarkName to be aware of such potential biases and risks of misuse. We note that any biases that might exist in the original resources gathered in this benchmark are unintentional and do not aim to cause harm.

\subsection*{Intended Use}
The \benchmarkName benchmark is intended to promote the development and the rigorous evaluation of language models for Bulgarian. We further believe that the benchmark will serve to examine the capabilities and the limitations of existing and emerging models on the challenging natural language understanding tasks in Bulgarian. Ideally, this could also lead to raising awareness of the potential risks associated with the use of such models developed for downstream tasks in Bulgarian.

\subsection*{Environmental Impact}
While \benchmarkName can stimulate the development of new machine learning models, it is worth noting that such models could require large computational resources for training, which contributes to global warming \cite{strubell-etal-2019-energy}. On the other hand, \benchmarkName is intended mainly for fine-tuning of pre-trained large language models, which requires considerably smaller computations. Additionally, we release the benchmark and the models on the HuggingFace Hub, which further reduces the environmental impact, as fine-tuning again is computationally costly, especially for larger models.

\bibliography{bibliography}
\clearpage

\appendix

\section*{Appendix}

\section{Model Hyper-Parameters and Training}
\label{appx:model:hyperparams}
Below, we first describe the values of some parameters that are across all models we experiment with, and then we discuss the values of some model-specific parameters:

\begin{itemize}
    \item All our models use the AdamW~\cite{loshchilov2018decoupled} optimizer with a weight decay of 1e-8, $\beta_1$ 0.9, $\beta_2$ 0.999, $\epsilon$ 1e-08, and are trained for five epochs with a batch size of 16 (gradient accumulation is applied when needed), and a maximum length of 512 tokens.
    \item We truncate longer input sequences token by token, if the input is formed from multiple sequences (see Section~\ref{appx:model:inputs}), i.e.,~pairs, we start from the longest one.
    \item All models use a warmup ration of 0.06 from the training data. We experiment with learning rate values \{2--5\}e-04 for base and distilled models, and \{1--3\}e-04 for XLM-R\textsubscript{Large}.
    \item The values of the hyper-parameters (including the number of training epochs) and the best checkpoints were selected on the development set. We use the target metric for each task as a checkpoint selection criterion.
    \item We trained our models on 5x Tesla T4 GPUs. Depending on the dataset size, the experiments took between 10 minutes, with the smaller datasets and models, and up 2 hours, for larger datasets. Training the XNLI model took 10 hours with base models, and 20 hours for large models.
    \item All models were trained with half precision (fp16) using the default PyTorch implementation.
    \item Table~\ref{tab:model:parameters} shows the models' size in terms of number of parameters.
    \item When evaluating the \emph{Token Classification Tasks} if the predicted sequence was shorter than the target one (i.e.,~not all inputs fit into 512 tokens), we added empty tags ('\emph{O}') until the target length was reached.
\end{itemize}


\begin{table}[ht!]
    \centering
    \begin{tabular}{lc}
        \toprule
         \bf{Model Name} & \bf{\#Params} \\
         \midrule
         XLM-R\textsubscript{large} & 560M \\
         XLM-R\textsubscript{base} & 278M \\
         SlavicBERT & 178 \\
         mBERT\textsubscript{base} & 178M \\
         Distil-mBERT & 135 \\
         MiniLM\textsubscript{L12} & 118M \\
         \bottomrule
    \end{tabular}
    \caption{Number of parameters in millions for each baseline pre-trained model included in the evaluation.}
    \label{tab:model:parameters}
\end{table}

\begin{table*}[th!]
    \centering
    \resizebox{\textwidth}{!}{%
    \setlength{\tabcolsep}{3.7pt}
        \begin{tabular}{llll}
        \toprule
             \bf{Task} & \bf{Input} & \bf{Output} & \bf{Loss} \\
        \midrule
             \bf{BSNLP} & [CLS] \textit{Document} [SEP] & BIO Tag & Per Token Cross Entropy \\
             \bf{Cinexo} & [CLS] \textit{User Comment} [SEP] & Rating (1--5) & Mean Squared Error \\
             \bf{CT21.T1} & [CLS] \textit{Tweet} [SEP] & Normal / Check-worthy & Binary Cross Entropy \\
             \bf{Cred.-N} & [CLS] \textit{Title} [SEP] \textit{News Article} [SEP] & Credible / Humorous & Binary Cross Entropy \\
             \bf{Exams} & [CLS] \textit{Context} [SEP] \textit{Question + Option} [SEP] & Option Ranking & Binary Cross Entropy \\
             \bf{Fake-N} & [CLS] \textit{Title} [SEP] \textit{News Article} [SEP] & Credible / Fake & Binary Cross Entropy \\
             \bf{PAN-X} & [CLS] \textit{Wikipedia sentence} [SEP] & BIO Tag & Per Token Cross Entropy \\
             \bf{U.Dep} & [CLS] \textit{Document} [SEP] & POS Tag & Per Token Cross Entropy \\
             \bf{XNLI} & [CLS] \textit{Hypothesis} [SEP] \textit{Premise} [SEP] & Entailment (3-way)  & Cross Entropy \\
        \bottomrule
        \end{tabular}
    }
    \caption{Input format for each task, the special tokens are replaced with the corresponding ones from the baseline model. Expected output, e.g.,~tag name, class, ranking, rating, etc. Finally, the optimization loss used for training.}
    \label{tab:input:output}
\end{table*}

\section{Model Input, Output and Loss}
\label{appx:model:inputs}

Table~\ref{tab:input:output} shows the inputs and the outputs for each model. We selected the formats based on  previous work~\cite{devlin2019bert,liu2019roberta} and the proposed formats on the (Super)GLUE benchmark~\cite{wang-etal-2019-glue,neurips2019_4496bf24}. For all tasks we introduce a projection layer on top of the pre-trained language model's representations. For classification tasks, the output maps to the number of classes, for regression, we project it to a single continuous value, for ranking, we obtain a probability distribution over two classes, for question answering, we rank each answer based on the log probability score, and finally, for token classification tasks, we apply the classification head on top of each token's representation. It is important to note that we use the BIO encoding for the NER tasks. We chose the loss function based on the target value. Finally, we replaced the special tokens with the corresponding ones from the baseline model.

\begin{table}[t]
    \centering
    \begin{tabular}{lr}
    \toprule
        \bf{Topic} & \bf{Examples}  \\
        \midrule
        Brexit & 598 \\
        Covid19 & 151 \\
        USElection2020 & 150 \\
        NordStream & 130 \\
        AsiaBibi & 94 \\
        Ryanair & 84 \\
        \midrule
        Total & 1,207 \\
        \bottomrule
    \end{tabular}
    \caption{Topic distribution in the \emph{BSNLP} dataset.}
    \label{tab:bsnlp:topics}
\end{table}

\begin{table}[t]
    \centering
    \begin{tabular}{lr}
        \toprule
         \bf{Subset} & \bf{\#Unique Movies}  \\
         \midrule
         Train & 257 \\
         Dev & 25 \\
         Test & 47 \\
         \midrule
         Total & 329 \\
         \bottomrule
    \end{tabular}
    \caption{Number of unique movies in each subset in the \emph{Cinexio} dataset that the users comment about.}
    \label{tab:cinexio:movies}
\end{table}

\begin{table}[t]
    \centering
    \begin{tabular}{lr}
        \toprule
         \bf{Subset} & \bf{\#Choices} \\
         \midrule
            Train 	& 3.88 \\
            Dev 	& 4.00 \\
            Test 	& 4.00  \\
        \bottomrule
    \end{tabular}
    \caption{Number of options per question in the \emph{EXAMS} dataset.}
    \label{tab:exams:choices}
\end{table}

\begin{figure}[t]
    \centering
    \includegraphics[width=\columnwidth]{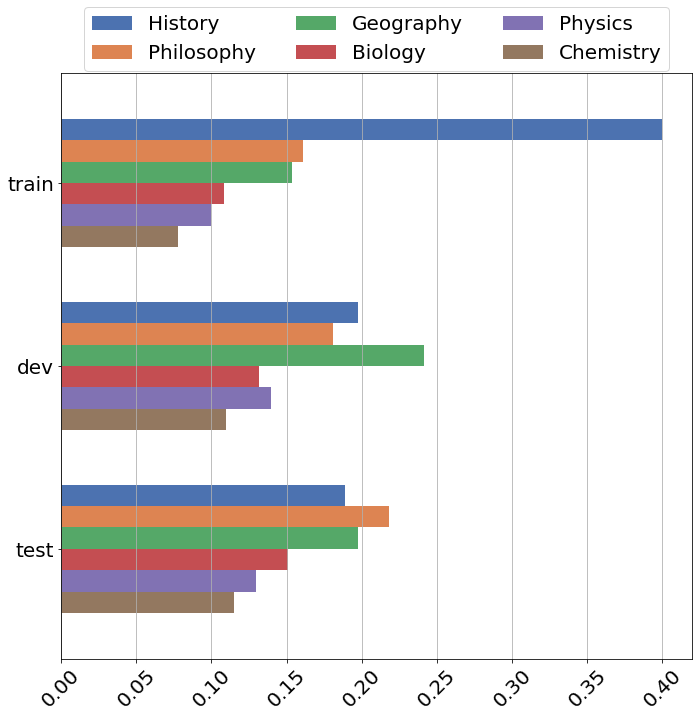}
    \caption{Subject distribution in the \emph{EXAMS} dataset.}
    \label{fig:exams:subjects}
\end{figure}

\begin{table}[t]
    \centering
      \begin{tabular}{lrrr}
      \toprule
      {} &  P/S Corr. &  Pearson &  Spearman \\
      \midrule
      XLM-R\textsubscript{Large}                      &           85.69 &     89.66 &      81.73 \\
      XLM-R\textsubscript{Base}                       &           84.40 &     87.91 &      80.90 \\
      SlavicBERT &           81.71 &     84.76 &      78.66 \\
      mBERT\textsubscript{Base}           &           82.07 &     85.22 &      78.92 \\
      MiniLM\textsubscript{L12} &           80.63 &     85.05 &      76.21 \\
      DistilBERT     &           80.32 &     83.55 &      77.09 \\
      \bottomrule
      \end{tabular}
      
  \caption{Fine-grained results for \textbf{Cinexio}.}
    \label{tab:fg:cinexio}
\end{table}

\begin{table*}[t]
    \centering
      \begin{tabular}{lrrrrrrrr}
      \toprule
      {} &  Avg. P &    P@1 &    P@3 &    P@5 &  P@10 &  P@20 &  P@50 &  R-Precision \\
      \midrule
      XLM-R\textsubscript{Large}                      &          69.45 & 100.00 & 100.00 & 100.00 & 90.00 & 90.00 & 70.00 &        59.21 \\
      XLM-R\textsubscript{Base}                       &          63.91 & 100.00 & 100.00 & 100.00 & 90.00 & 75.00 & 68.00 &        57.89 \\
      SlavicBERT &          62.70 & 100.00 & 100.00 & 100.00 & 80.00 & 70.00 & 64.00 &        61.84 \\
      mBERT\textsubscript{Base}           &          64.79 & 100.00 & 100.00 &  80.00 & 90.00 & 85.00 & 72.00 &        60.53 \\
      MiniLM\textsubscript{L12} &          57.37 & 100.00 &  66.67 &  60.00 & 70.00 & 75.00 & 66.00 &        59.21 \\
      DistilBERT     &          65.15 & 100.00 & 100.00 &  80.00 & 90.00 & 90.00 & 68.00 &        56.58 \\
      \bottomrule
      \end{tabular}
      
  \caption{Fine-grained results for \textbf{CLEF-2021 CheckThat!, Task 1A (CT21.T1)}.}
    \label{tab:fg:clef}
\end{table*}

\begin{table}[t]
    \centering
      \begin{tabular}{lrrr}
      \toprule
      {} & \multicolumn{3}{c}{Humorous} \\
      {} & F1 & P & R \\
      \midrule
      XLM-R\textsubscript{Large}                      &    79.73 &     86.52 &  73.93 \\
      XLM-R\textsubscript{Base}                       &    75.74 &     77.53 &  74.04 \\
      SlavicBERT &    72.01 &     88.97 &  60.48 \\
      mBERT\textsubscript{Base}           &    69.17 &     88.91 &  56.60 \\
      MiniLM\textsubscript{L12} &    75.41 &     76.55 &  74.31 \\
      DistilBERT     &    67.05 &     83.27 &  56.11 \\
      \bottomrule
      \end{tabular}
      
  \caption{Fine-grained results for \textbf{Credible News (Cred.-N)}.}
    \label{tab:fg:crediblenews}
\end{table}

\begin{table}[t]
    \centering
      \begin{tabular}{lrrr}
      \toprule
      {} & \multicolumn{3}{c}{Fake} \\
      {} & F1 & P & R \\
      \midrule
      XLM-R\textsubscript{Large}                      &    70.31 &     68.20 &  72.55 \\
      XLM-R\textsubscript{Base}                       &    66.82 &     61.22 &  73.53 \\
      SlavicBERT &    67.28 &     63.09 &  72.06 \\
      mBERT\textsubscript{Base}           &    65.65 &     68.25 &  63.24 \\
      MiniLM\textsubscript{L12} &    64.33 &     58.10 &  72.06 \\
      DistilBERT     &    65.66 &     67.18 &  64.22 \\
      \bottomrule
      \end{tabular}
      
  \caption{Fine-grained results for \textbf{Fake News (Fake-N)}.}
    \label{tab:fg:fakenews}
\end{table}

\begin{table*}[t]
    \centering
      \begin{tabular}{llrrrrrr}
      \toprule
          &        &  XLM-R\textsubscript{Large} &  XLM-R\textsubscript{Base} &  SlavicBERT &  mBERT\textsubscript{Base} &  MiniLM\textsubscript{L12} &  DistilBERT \\
      \midrule
      \multirow{3}{*}{Overall} & F1 &              63.81 &             62.47 &                                   65.28 &                         56.13 &                                   59.70 &                               52.82 \\
      {} & P &               87.22 &             85.69 &                                   84.36 &                         83.73 &                                   81.55 &                               76.70 \\
      {} & R &               50.30 &             49.15 &                                   53.24 &                         42.21 &                                   47.08 &                               40.28 \\
      \midrule
      \multirow{3}{*}{EVT} & F1 &               4.52 &              5.73 &                                   10.56 &                          3.16 &                                    0.20 &                                1.36 \\
          & P &              62.16 &             45.45 &                                   49.15 &                         51.61 &                                   10.00 &                               13.46 \\
          & R &               2.34 &              3.06 &                                    5.91 &                          1.63 &                                    0.10 &                                0.71 \\
      \midrule
      \multirow{3}{*}{LOC} & F1 &              74.95 &             73.89 &                                   79.25 &                         68.46 &                                   71.45 &                               64.94 \\
          & P &              95.06 &             92.81 &                                   92.55 &                         92.02 &                                   91.00 &                               86.73 \\
          & R &              61.86 &             61.38 &                                   69.29 &                         54.50 &                                   58.82 &                               51.90 \\
      \midrule
      \multirow{3}{*}{ORG} & F1 &              55.50 &             53.05 &                                   57.22 &                         49.03 &                                   50.02 &                               42.59 \\
          & P &              74.00 &             71.31 &                                   71.31 &                         68.28 &                                   59.06 &                               56.41 \\
          & R &              44.40 &             42.23 &                                   47.77 &                         38.25 &                                   43.37 &                               34.22 \\
      \midrule
      \multirow{3}{*}{PER} & F1 &              72.83 &             71.58 &                                   72.55 &                         62.84 &                                   68.83 &                               60.92 \\
          & P &              97.80 &             97.15 &                                   97.57 &                         97.27 &                                   94.05 &                               91.60 \\
          & R &              58.02 &             56.66 &                                   57.74 &                         46.41 &                                   54.28 &                               45.64 \\
      \midrule
      \multirow{3}{*}{PRO} & F1 &              40.91 &             38.56 &                                   36.42 &                         34.36 &                                   35.39 &                               32.55 \\
          & P &              40.91 &             40.30 &                                   34.84 &                         35.84 &                                   40.83 &                               33.54 \\
          & R &              40.91 &             36.96 &                                   38.14 &                         33.00 &                                   31.23 &                               31.62 \\
      \bottomrule
      \end{tabular}
      
  \caption{Fine-grained results for \textbf{BSNLP}.}
    \label{tab:fg:bsnlp}
\end{table*}

\begin{table*}[t]
    \centering
      \begin{tabular}{llrrrrrr}
      \toprule
          &        &  XLM-R\textsubscript{Large} &  XLM-R\textsubscript{Base} &  SlavicBERT &  mBERT\textsubscript{Base} &  MiniLM\textsubscript{L12} &  DistilBERT \\
      \midrule
      \multirow{3}{*}{Overall} & F1 &              92.96 &             91.18 &                                   92.36 &                         92.11 &                                   90.26 &                               90.82 \\
      {} & P &               92.37 &             90.77 &                                   91.85 &                         91.70 &                                   89.63 &                               90.40 \\
      {} & R &               93.55 &             91.59 &                                   92.88 &                         92.52 &                                   90.91 &                               91.24 \\
      \midrule
      \multirow{3}{*}{LOC} & F1 &              95.21 &             93.66 &                                   94.97 &                         94.43 &                                   93.37 &                               93.88 \\
          & P &              95.03 &             92.85 &                                   94.53 &                         93.80 &                                   93.07 &                               93.57 \\
          & R &              95.39 &             94.50 &                                   95.41 &                         95.08 &                                   93.67 &                               94.19 \\
      \midrule
      \multirow{3}{*}{ORG} & F1 &              86.33 &             83.50 &                                   84.75 &                         84.81 &                                   81.82 &                               82.80 \\
          & P &              85.35 &             84.15 &                                   84.80 &                         84.81 &                                   80.86 &                               82.57 \\
          & R &              87.34 &             82.85 &                                   84.70 &                         84.81 &                                   82.81 &                               83.04 \\
      \midrule
      \multirow{3}{*}{PER} & F1 &              95.07 &             93.67 &                                   94.69 &                         94.60 &                                   92.61 &                               92.82 \\
          & P &              94.25 &             92.92 &                                   93.56 &                         94.17 &                                   91.80 &                               92.06 \\
          & R &              95.90 &             94.43 &                                   95.84 &                         95.03 &                                   93.43 &                               93.59 \\
      \bottomrule
      \end{tabular}
      
  \caption{Fine-grained results for \textbf{PAN-X (WikiAnn)}.}
    \label{tab:fg:wikiann}
\end{table*}

\begin{table*}[t]
    \centering
    \resizebox{0.87\textwidth}{!}{%
    
    \setlength{\tabcolsep}{3.7pt}
      \begin{tabular}{llrrrrrr}
      \toprule
         &        &  XLM-R\textsubscript{Large} &  XLM-R\textsubscript{Base} &  SlavicBERT &  mBERT\textsubscript{Base} &  MiniLM\textsubscript{L12} &  DistilBERT \\
      \midrule
      \multirow{3}{*}{Overall} & F1 &              99.30 &             99.23 &                                   99.06 &                         98.99 &                                   98.91 &                               98.58 \\
      {} & P &               99.32 &             99.24 &                                   99.08 &                         99.00 &                                   98.92 &                               98.59 \\
      {} & R &               99.29 &             99.22 &                                   99.04 &                         98.98 &                                   98.91 &                               98.57 \\
      \midrule
      \multirow{3}{*}{ART} & F1 &              99.22 &             99.23 &                                   98.22 &                         97.97 &                                   97.46 &                               96.46 \\
         & P &              99.48 &             98.48 &                                   96.98 &                         96.50 &                                   96.00 &                               94.55 \\
         & R &              98.97 &            100.00 &                                   99.48 &                         99.48 &                                   98.97 &                               98.45 \\
      \midrule
      \multirow{3}{*}{CONJ} & F1 &              99.76 &             99.68 &                                   99.68 &                         99.51 &                                   99.11 &                               99.27 \\
         & P &              99.68 &             99.68 &                                   99.52 &                         99.51 &                                   99.19 &                               99.35 \\
         & R &              99.84 &             99.68 &                                   99.84 &                         99.51 &                                   99.03 &                               99.19 \\
      \midrule
      \multirow{3}{*}{DJ} & F1 &              99.19 &             98.77 &                                   98.69 &                         98.42 &                                   98.23 &                               97.10 \\
         & P &              99.08 &             98.54 &                                   98.47 &                         98.61 &                                   97.93 &                               97.44 \\
         & R &              99.31 &             99.00 &                                   98.92 &                         98.23 &                                   98.54 &                               96.77 \\
      \midrule
      \multirow{3}{*}{DP} & F1 &              99.96 &             99.91 &                                   99.89 &                         99.93 &                                   99.93 &                               99.82 \\
         & P &              99.96 &             99.96 &                                   99.91 &                         99.96 &                                   99.96 &                               99.87 \\
         & R &              99.96 &             99.87 &                                   99.87 &                         99.91 &                                   99.91 &                               99.78 \\
      \midrule
      \multirow{3}{*}{DV} & F1 &              99.35 &             98.52 &                                   97.94 &                         98.93 &                                   97.77 &                               97.39 \\
         & P &              99.18 &             99.01 &                                   98.83 &                         99.18 &                                   98.67 &                               97.23 \\
         & R &              99.51 &             98.04 &                                   97.05 &                         98.69 &                                   96.89 &                               97.55 \\
      \midrule
      \multirow{3}{*}{ERB} & F1 &              99.17 &             99.14 &                                   98.81 &                         98.66 &                                   98.72 &                               98.24 \\
         & P &              99.52 &             99.46 &                                   99.28 &                         98.93 &                                   99.10 &                               98.51 \\
         & R &              98.81 &             98.81 &                                   98.34 &                         98.40 &                                   98.34 &                               97.98 \\
      \midrule
      \multirow{3}{*}{ET} & F1 &              97.75 &             97.74 &                                   96.81 &                         96.42 &                                   96.64 &                               95.67 \\
         & P &              98.49 &             99.23 &                                   97.73 &                         97.71 &                                   97.00 &                               96.95 \\
         & R &              97.03 &             96.28 &                                   95.91 &                         95.17 &                                   96.28 &                               94.42 \\
      \midrule
      \multirow{3}{*}{NTJ} & F1 &              96.97 &             96.97 &                                   96.97 &                        100.00 &                                   80.00 &                               90.32 \\
         & P &             100.00 &            100.00 &                                  100.00 &                        100.00 &                                   77.78 &                              100.00 \\
         & R &              94.12 &             94.12 &                                   94.12 &                        100.00 &                                   82.35 &                               82.35 \\
      \midrule
      \multirow{3}{*}{OUN} & F1 &              99.39 &             99.42 &                                   99.42 &                         99.16 &                                   99.26 &                               98.85 \\
         & P &              99.29 &             99.61 &                                   99.41 &                         99.29 &                                   99.23 &                               98.85 \\
         & R &              99.50 &             99.23 &                                   99.44 &                         99.02 &                                   99.29 &                               98.85 \\
      \midrule
      \multirow{3}{*}{RON} & F1 &              99.56 &             99.46 &                                   98.86 &                         98.86 &                                   98.91 &                               98.54 \\
         & P &              99.56 &             99.14 &                                   98.80 &                         98.49 &                                   99.02 &                               98.06 \\
         & R &              99.56 &             99.78 &                                   98.91 &                         99.24 &                                   98.80 &                               99.02 \\
      \midrule
      \multirow{3}{*}{ROPN} & F1 &              97.52 &             97.87 &                                   97.55 &                         97.62 &                                   97.31 &                               97.31 \\
         & P &              97.83 &             96.69 &                                   96.95 &                         96.96 &                                   96.79 &                               96.94 \\
         & R &              97.22 &             99.07 &                                   98.15 &                         98.30 &                                   97.84 &                               97.69 \\
      \midrule
      \multirow{3}{*}{UM} & F1 &              96.76 &             96.54 &                                   96.30 &                         95.61 &                                   96.06 &                               96.04 \\
         & P &              95.87 &             95.43 &                                   95.41 &                         94.52 &                                   95.39 &                               95.81 \\
         & R &              97.66 &             97.66 &                                   97.20 &                         96.73 &                                   96.73 &                               96.26 \\
      \midrule
      \multirow{3}{*}{UNCT} & F1 &              99.98 &             99.98 &                                   99.98 &                         99.98 &                                   99.98 &                               99.98 \\
         & P &              99.95 &             99.95 &                                   99.95 &                         99.95 &                                   99.95 &                               99.95 \\
         & R &             100.00 &            100.00 &                                  100.00 &                        100.00 &                                  100.00 &                              100.00 \\
      \midrule
      \multirow{3}{*}{UX} & F1 &              97.80 &             97.80 &                                   97.55 &                         97.56 &                                   97.61 &                               97.19 \\
         & P &              97.80 &             97.68 &                                   97.55 &                         97.44 &                                   97.56 &                               97.07 \\
         & R &              97.80 &             97.92 &                                   97.55 &                         97.67 &                                   97.67 &                               97.31 \\
      \bottomrule
      \end{tabular}
      }
      
  \caption{Fine-grained results for \textbf{Universal Dependencies (U. Dep).}}
    \label{tab:fg:udep}
\end{table*}

\section{Additional Task Details}
\label{appx:task:details}

In this section, we summarize some additional characteristics for each task in the \benchmarkName benchmark.

\begin{figure*}[t]
     \centering
     \begin{subfigure}[b]{0.3\textwidth}
         \centering
         \includegraphics[width=\textwidth]{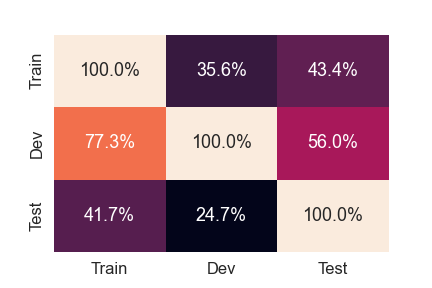}
         \caption{BSNLP}
         \label{fig:wo:bsnlp}
     \end{subfigure}
     \hfill
     \begin{subfigure}[b]{0.3\textwidth}
         \centering
         \includegraphics[width=\textwidth]{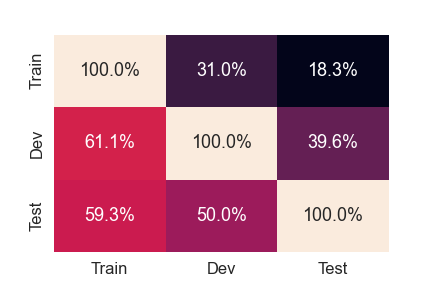}
         \caption{Cinexio}
         \label{fig:wo:cinexio}
     \end{subfigure}
     \hfill
     \begin{subfigure}[b]{0.3\textwidth}
         \centering
         \includegraphics[width=\textwidth]{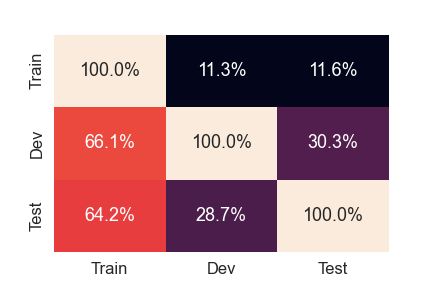}
         \caption{CT21.T1}
         \label{fig:wo:ct21t1}
     \end{subfigure}
     \begin{subfigure}[b]{0.3\textwidth}
         \centering
         \includegraphics[width=\textwidth]{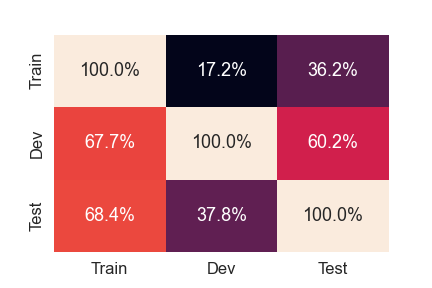}
         \caption{Cred.-N}
         \label{fig:wo:credn}
     \end{subfigure}
     \hfill
     \begin{subfigure}[b]{0.3\textwidth}
         \centering
         \includegraphics[width=\textwidth]{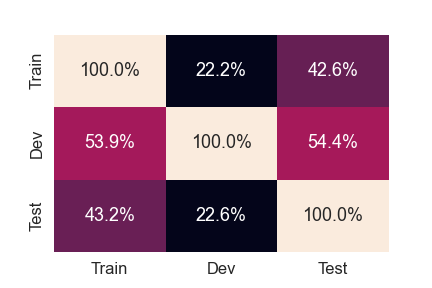}
         \caption{EXAMS}
         \label{fig:wo:exams}
     \end{subfigure}
     \hfill
     \begin{subfigure}[b]{0.3\textwidth}
         \centering
         \includegraphics[width=\textwidth]{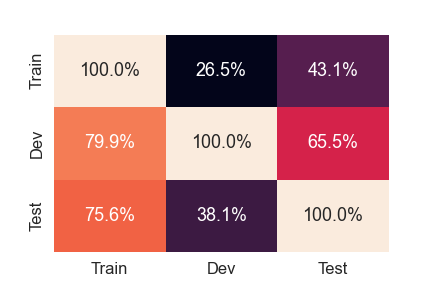}
         \caption{Fake-N}
         \label{fig:wo:faken}
     \end{subfigure}
     \begin{subfigure}[b]{0.3\textwidth}
         \centering
         \includegraphics[width=\textwidth]{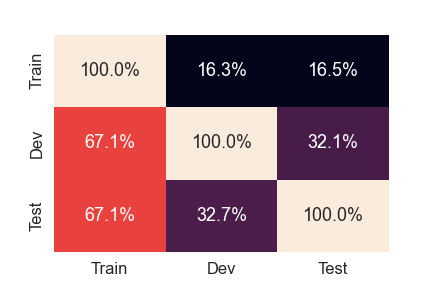}
         \caption{U.Dep}
         \label{fig:wo:udep}
     \end{subfigure}
     \hfill
     \begin{subfigure}[b]{0.3\textwidth}
         \centering
         \includegraphics[width=\textwidth]{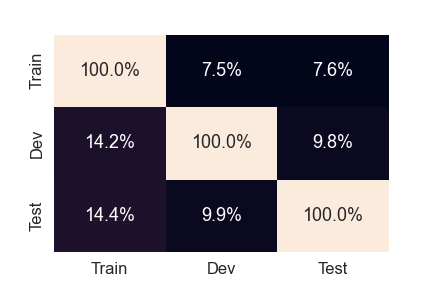}
         \caption{PAN-X/WikiANN}
         \label{fig:wo:panx}
     \end{subfigure}
     \hfill
     \begin{subfigure}[b]{0.3\textwidth}
         \centering
         \includegraphics[width=\textwidth]{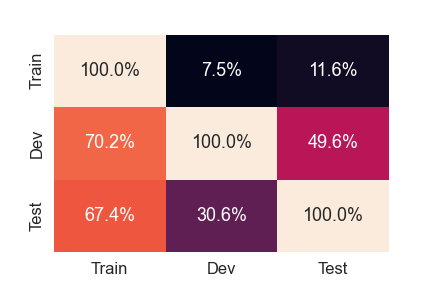}
         \caption{XNLI}
         \label{fig:wo:xnli}
     \end{subfigure}
        \caption{Per-task vocabulary overlap. Calculated as the number of common words in the row and the column divided by the total number of unique words in the row.}
        \label{fig:wo:plots}
\end{figure*}

Figure~\ref{fig:wo:plots} shows some statistics about the word overlap between subsets. To calculate the statistics, we split the texts into words using the NLTK~\citet{bird2009natural} tokenizer. After that, we take the number of unique words in each subset and we take the union of all common words between the first and the second subset, we then compare and divide them by the size of the super set obtained by combining the two. We see that most of the datasets have high overlap between the training and  the development/testing set. This is expected as the training sets are often significantly larger compared to the other subsets, and also as we did not filter out the stop words, which cover a big part of the word tokens. 

Interestingly, the only exception is the PAN-X dataset. We attribute this to the text snippets being short, designed to contain named entities, and being extracted from different Wikipedia articles.

\begin{figure*}[t]
     \centering
     \begin{subfigure}[b]{0.3\textwidth}
         \centering
         \includegraphics[width=\textwidth]{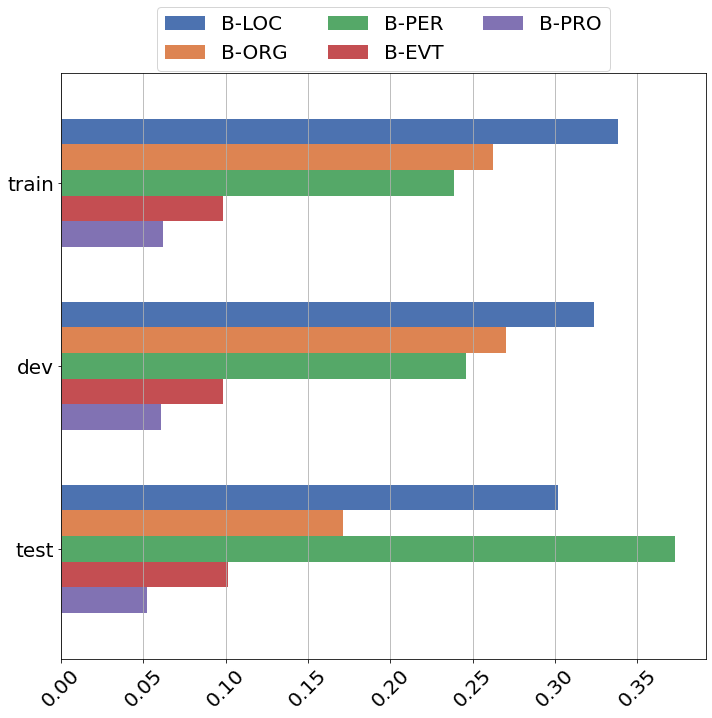}
         \caption{BSNLP}
         \label{fig:labels:bsnlp}
     \end{subfigure}
     \hfill
     \begin{subfigure}[b]{0.3\textwidth}
         \centering
         \includegraphics[width=\textwidth]{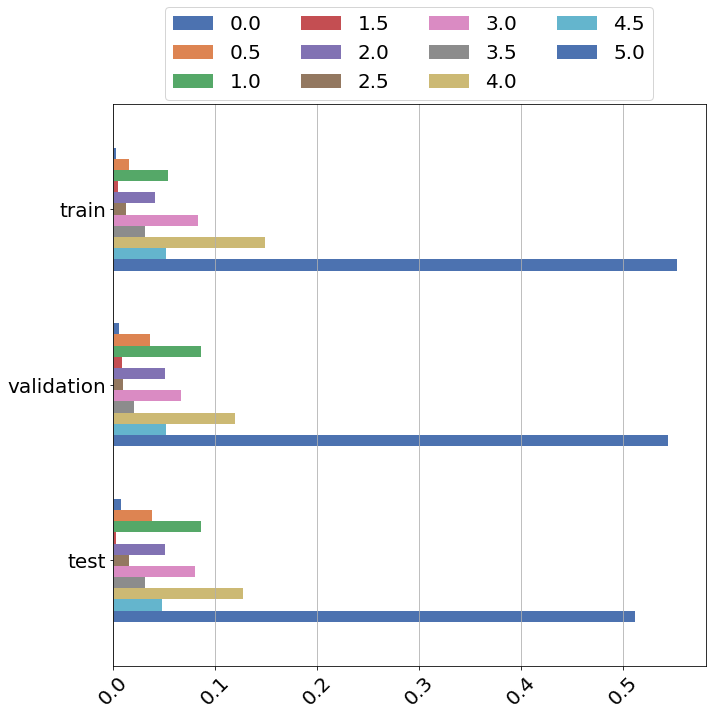}
         \caption{Cinexio}
         \label{fig:labels:cinexio}
     \end{subfigure}
     \hfill
     \begin{subfigure}[b]{0.3\textwidth}
         \centering
         \includegraphics[width=\textwidth]{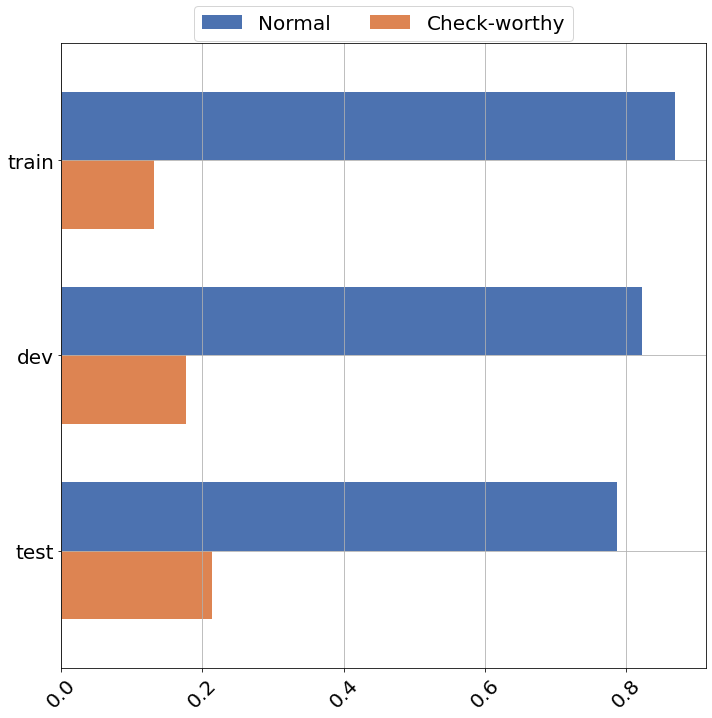}
         \caption{CT21.T1}
         \label{fig:labels:ct21t1}
     \end{subfigure}
     \begin{subfigure}[b]{0.3\textwidth}
         \centering
         \includegraphics[width=\textwidth]{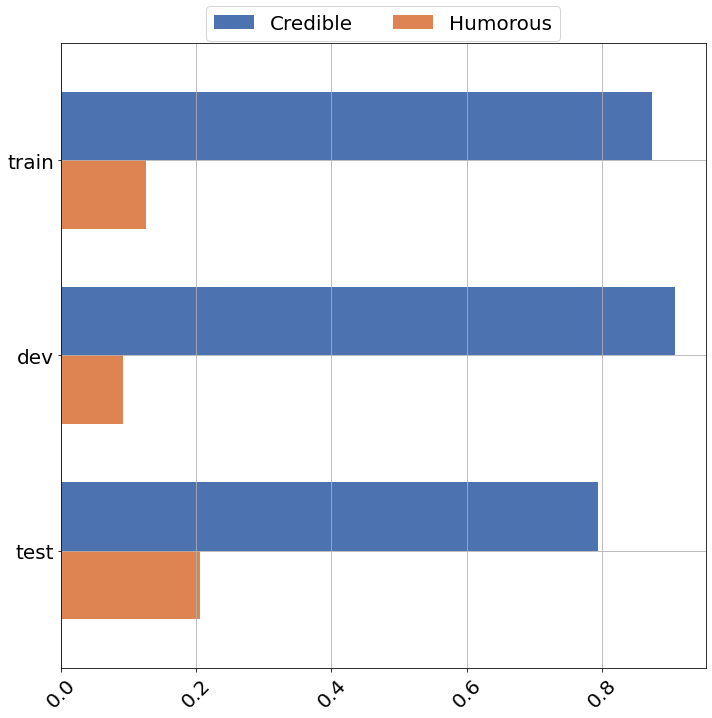}
         \caption{Cred.-N}
         \label{fig:labels:credn}
     \end{subfigure}
     \hfill
     \begin{subfigure}[b]{0.3\textwidth}
         \centering
         \includegraphics[width=\textwidth]{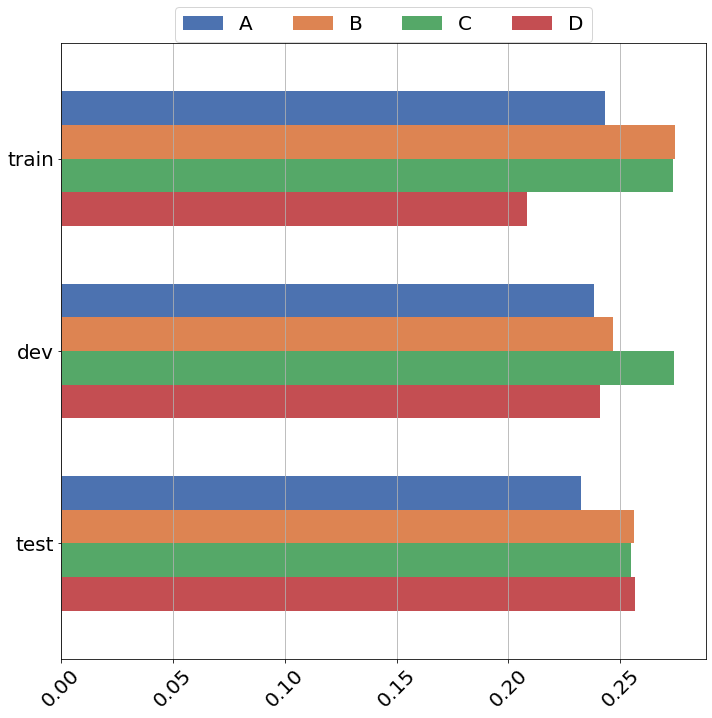}
         \caption{EXAMS}
         \label{fig:labels:exams}
     \end{subfigure}
     \hfill
     \begin{subfigure}[b]{0.3\textwidth}
         \centering
         \includegraphics[width=\textwidth]{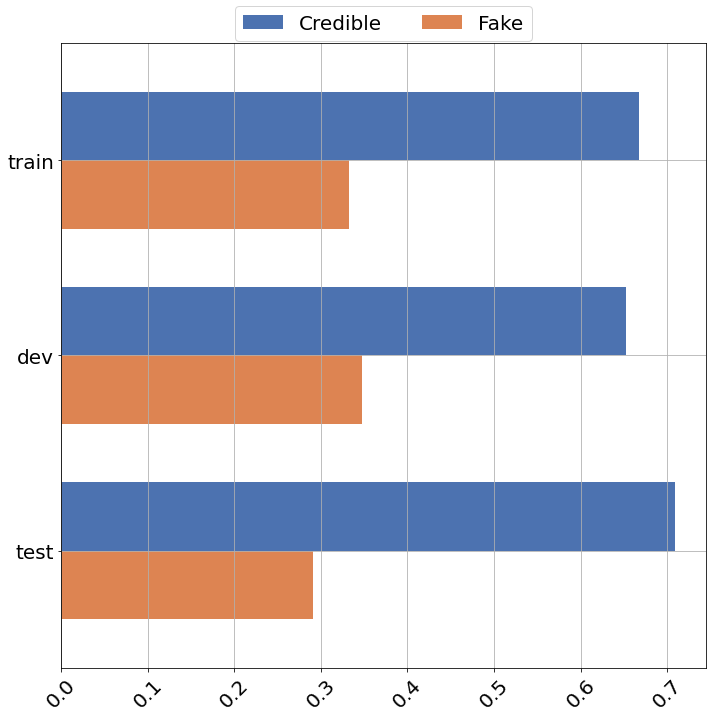}
         \caption{Fake-N}
         \label{fig:labels:faken}
     \end{subfigure}
     \begin{subfigure}[b]{0.3\textwidth}
         \centering
         \includegraphics[width=\textwidth]{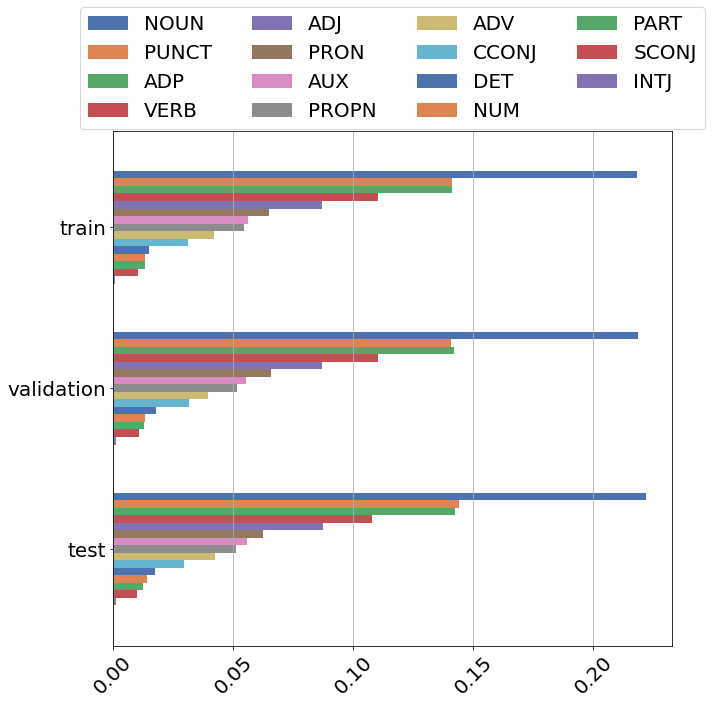}
         \caption{U.Dep}
         \label{fig:labels:udep}
     \end{subfigure}
     \hfill
     \begin{subfigure}[b]{0.3\textwidth}
         \centering
         \includegraphics[width=\textwidth]{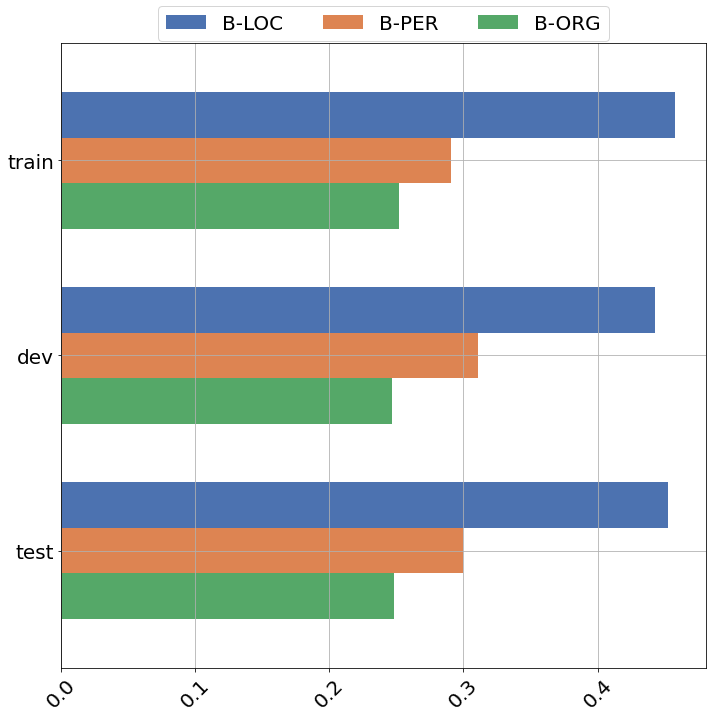}
         \caption{PAN-X/WikiANN}
         \label{fig:labels:panx}
     \end{subfigure}
     \hfill
     \begin{subfigure}[b]{0.3\textwidth}
         \centering
         \includegraphics[width=\textwidth]{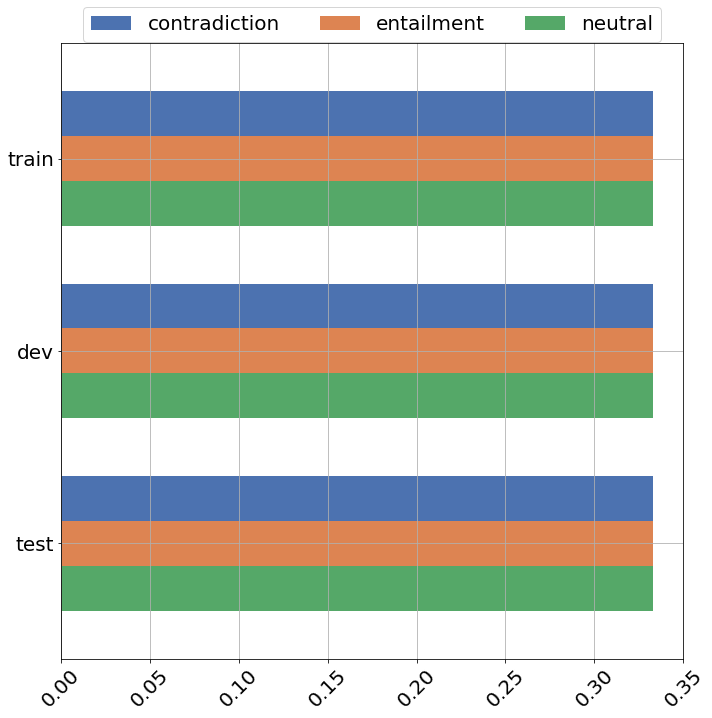}
         \caption{XNLI}
         \label{fig:labels:xnli}
     \end{subfigure}
        \caption{Per-task label distribution. We remove the empty tags when we plot for the NER tasks.}
        \label{fig:labels:plots}
\end{figure*}

Figure~\ref{fig:labels:plots} shows the per task label distribution. We see that most of the tasks maintain similar distributions across labels, expect for BSNLP, where we have less ORG and PRO tags and more PER.

\paragraph{BSNLP} We can see in Table~\ref{tab:bsnlp:topics} that the most represented topic is Brexit with 600 examples (4x compared to the second topic), followed by COVID-19, US Elections 2020, and Nord Stream, each covering well above 100 examples. The other two topics, AsiaBibi and Ryanair, have less than 100 examples.

\paragraph{Cinexio} Table~\ref{tab:cinexio:movies} shows the number of movies in the Cinexio dataset.  Each movie received 29.9 comments on average.

\paragraph{Cred.-N.} We used a custom crawler, \emph{BeautifulSoup} to parse the HTML, and per-site CSS selectors to extract the articles' text. The crawler was based on simple rules that collect and follow the links to articles in each starting page we pass. Our starting points are pages that contain all articles sorted by their publication date and paginated. Finally, we remove all HTML tags, images and information about the authors and the sources, retaining only the plain text. We annotated the articles as credible or humorous based on the label for their website. More details about the dataset and the pre-processing are given in Sections~\ref{sec:tasks} and \ref{sec:data:preparation}.

\paragraph{High School Examinations (EXAMS)} Figure~\ref{fig:exams:subjects} shows the average number of options per question for each subset in the datasets. Both the \emph{Dev} and \emph{Test} subset have four options, but \emph{Train} contains questions with three answers coming from online history exams collected from~\citet{hardalov-etal-2019-beyond}. These examples also affect the subject distribution for the training set.

\paragraph{Fake News (Fake-N.)} Here, we report the number of unique and common domains:

\begin{itemize}[nosep]
    \item  Train vs Dev
    \begin{itemize}
        \item \#Common Domains: 106
        \item Only in train: 239
        \item Only in dev: 13
    \end{itemize}
    \item Train vs. Test
    \begin{itemize}
        \item \#Common Domains: 162
        \item Only in train: 183
        \item Only in test: 46
    \end{itemize}
    \item Dev vs. Test
    \begin{itemize}
        \item \#Common Domains: 90
        \item Only in dev: 29
        \item Only in test: 118
    \end{itemize}

\end{itemize}

\section{Fine-Grained Results} 
\label{appx:finegrained:results}
Here, we present the fine-grained results per task. Grouped by the task types from Table~\ref{tab:tasks_dataset}, we include the following tables: (\emph{i})~\emph{Regression / Ranking} -- 
In Table~\ref{tab:fg:cinexio}, we present the Spearman and the Pearson correlation values for the \emph{Cinexio} task.
Table~\ref{tab:fg:clef} shows the metrics for \emph{Ct21.T1}, including P@K and R-Precision; (\emph{ii})~\emph{Classification Tasks} -- for the binary classification tasks \emph{Cred.-N} and \emph{Fake-N} we include the \emph{Precision} and \emph{Recall} for the target class, i.e.,~\emph{Humorous}, and \emph{Fake} respectively, and finally (\emph{iii})~~\emph{Token Classification} -- Tables~\ref{tab:fg:bsnlp}, \ref{tab:fg:wikiann}, and \ref{tab:fg:udep} include per token type P, R, and F1.

 We did not include tables for \emph{EXAMS} and \emph{XNLI} as their target evaluation measure is \emph{Accuracy}, and thus they are only coarse-grained.


\begin{table*}[th!]
\centering \footnotesize
{
\renewcommand{\arraystretch}{1.1}
\begin{tabular}{p{0.005\textwidth}p{0.93\textwidth}}

 \toprule
 \parbox[t]{1mm}{\multirow{2}{*}{\rotatebox[origin=c]{90}{{\textbf{BSNLP}}}}} &
{\textbf{Document}: \textit{... The chancellor of \{Germany\}\textsuperscript{LOC} \{Angela Merkel\}\textsuperscript{PER} and the president of \{Russia\}\textsuperscript{LOC} \{Vladimir Putin\}\textsuperscript{PER} discussed over the phone the implementation of the project ``\{Nord Stream - 2\}\textsuperscript{PRO}'' ... Earlier the company ``\{Nord Stream\}\textsuperscript{ORG}'' which leads the construction ...}} \\
& \textbf{Possible Tags}: \underline{Person (PER)}, \underline{Organization (ORG)}, \underline{Location (LOC)}, \underline{Product (PRO)}, \underline{Event (EVT )} \\

\midrule
\parbox[t]{1mm}{\multirow{2}{*}{\rotatebox[origin=c]{90}{{\textbf{Cinexio}}}}} &
{\textbf{User Review}: \textit{Five stars are not enough - it deserves at least that many :)}} \\
& \textbf{Rating}: \underline{5.0} \\[1em]

\midrule
\parbox[t]{1mm}{\multirow{3}{*}{\rotatebox[origin=c]{90}{{\textbf{Cred.-N}}}}} & \textbf{Body:} \textit{Today is the deadline for Bulgarians living abroad to submit an application for opening a polling station for the upcoming referendum on January 27. According to a decision of the Central Election Commission (CEC), polling stations can be opened in the embassies and consulates of the country. However, for this purpose, at least 20 applications are needed from those who wish...} \\
& \textbf{Title}: \textit{Today is the deadline for submitting applications to open sections abroad for the referendum} \\
& \textbf{Correct Label}: \underline{Credible}\\

\midrule
\parbox[t]{1mm}{\multirow{2}{*}{\rotatebox[origin=c]{90}{{\textbf{CT21.T1}}}}} &
\textbf{Tweet:} \textit{According to research, \#COVID19 survives up to 3 hours in aerosols in the air, up to 24 hours on paper and about 2-3 days on a steel or plastic surface. [URL]}\\
& \textbf{Check-worthy}: \underline{Yes}\\[1em]

\midrule
\parbox[t]{1mm}{\multirow{2}{*}{\rotatebox[origin=c]{90}{{\textbf{EXAMS}}}}} &
\textbf{Paragraph:} \textit{In the autumn of 917 he sent an army ... to invade Serbia and punish Gojniković for his treachery. The Bulgarian ruler again sends Theodore Sigritsa and Marmais, but this time they are defeated... which forces Simeon to conclude a truce with Byzantium...} \quad \textbf{Subject}: \textit{History} \\
& \textbf{Question:} \\
& \textit{Which generals led Simeon's punitive campaign against the emerging Serbian danger in 917?} \\
& \textbf{Candidate answers:} \\
&
\textit{(\texttt{A}) \underline{Theodor Sigritsa and Marmais}},
\textit{(\texttt{B}) Cracra and Alusian},
\textit{(\texttt{C}) Ivac and Nikulitsa},
\textit{(\texttt{D}) Knin, Imnicus and Izvoklius} \\

\midrule
\parbox[t]{1mm}{\multirow{2}{*}{\rotatebox[origin=c]{90}{{\textbf{Fake.-N}}}}} &
\textbf{Body:} \textit{The researcher of Bulgarian prophets Hristo Radev reveals predictions of the Slava Sevryukova phenomenon in an interview for ``Bulgaria Today'' a person in whom the spirit of a bright biblical hero has been reborn. He means David. According to the clairvoyant, this Bulgarian will play a very important role in the future of the country. I hope this president is the person in question! Rumen Radev jumped out of nowhere, just like the biblical David...} \\
& \textbf{Title}: \textit{Petel.bg - news - ``Bulgaria today'': Slava Sevryukova's lost prophecy about Bulgaria was dug up! It is coming true before our eyes} \\
& \textbf{Correct Label}: \underline{Fake}\\

\midrule
\parbox[t]{1mm}{\multirow{2}{*}{\rotatebox[origin=c]{90}{{\textbf{PAN-X}}}}} &
\textbf{Sentence:} \textit{The species is distributed in \{Burundi\}\textsuperscript{LOC}, \{Democratic Republic of Congo\}\textsuperscript{LOC}, \{Zambia\}\textsuperscript{LOC} and \{ Tanzania\}\textsuperscript{LOC}.} \\
& \textbf{Possible Tags}: \underline{Person (PER)}, \underline{Organization (ORG)}, \underline{Location (LOC)} \\[1em]

\midrule
\parbox[t]{1mm}{\multirow{1}{*}{\rotatebox[origin=c]{90}{{\textbf{U.Dep}}}}} &
\textbf{Sentence:} \textit{In the discussion, I guess, important questions will be discussed.} \\
& \textbf{Possible Tags}: \\
& \underline{NOUN}, \underline{PUNCT}, \underline{ADP}, \underline{VERB}, \underline{ADJ}, \underline{PRON}, \underline{AUX}, \underline{PROPN}, \underline{ADV}, \underline{CCONJ}, \underline{DET}, \underline{NUM}, \underline{PART}, \underline{SCONJ}, \underline{INTJ} \\[1em]

\midrule
\parbox[t]{1mm}{\multirow{2}{*}{\rotatebox[origin=c]{90}{{\textbf{XNLI}}}}} &
\textbf{Text:} \textit{And he said, Mother, I am at home. He called his mother as soon as the school bus dropped him off.}\\
& \textbf{Hypothesis:} \textit{He called his mother as soon as the school bus dropped him off.} \\
& \textbf{Entailment:} \underline{Neutral}\\[0.5em]

\bottomrule
\end{tabular}
}
\caption{English translations of the examples shown in Table~\ref{tab:pertask:examples}.}
    \label{tab:appx:en:pertask:examples}
\end{table*}

\end{document}